\documentclass[10pt,twocolumn,letterpaper]{article}
\usepackage{cvpr}
\usepackage{times}
\usepackage{epsfig}
\usepackage{graphicx}
\usepackage{amsmath}
\usepackage{amssymb}
\usepackage[utf8]{inputenc}
\usepackage{amsthm}
\usepackage{etoolbox}
\usepackage{xspace}
\usepackage{array,multirow}
\usepackage{boldline}
\usepackage[titlenumbered,ruled ]{algorithm2e}
\usepackage[symbol]{footmisc}
\usepackage{tablefootnote}
\usepackage[flushleft]{threeparttable}
\usepackage{amsmath}
\usepackage{breqn}
\usepackage{tabularx}
\usepackage{booktabs}
\usepackage{multirow}
\usepackage{float}
\usepackage[pagebackref=true,breaklinks=true,letterpaper=true,colorlinks,bookmarks=false]{hyperref}
\usepackage[capitalise]{cleveref}
\usepackage{multicol}

\cvprfinalcopy 

\def\cvprPaperID{4926} 

\makeatletter	
	\newcommand{\removelatexerror}{\let\@latex@error\@gobble}
\makeatother
  

\newcommand{\MyMapTemplatePrefixc}[4]{\expandafter#1\csname#3#4\endcsname{#2{#4}}} 
\forcsvlist{\MyMapTemplatePrefixc {\def} {\mathcal}{c}} {A,B,C,D,E,F,G,H,I,J,K,L,M,N,O,P,Q,R,S,T,U,V,W,X,Y,Z}  
\newcommand{\MyMapTemplatePrefixtb}[5]{\expandafter#1\csname#4#5\endcsname{#2{#3{#5}}}} 
\forcsvlist{\MyMapTemplatePrefixtb {\def} {\tilde}{\mathbf}{t}} {A,B,C,D,E,F,G,H,I,J,K,L,M,N,O,P,Q,R,S,T,U,V,W,X,Y,Z}  
\forcsvlist{\MyMapTemplatePrefixtb {\def} {\tilde}{\mathbf}{t}} {0,1,a,b,c,d,e,f,g,h,i,j,k,l,m,n,o,p,q,r,s,u,v,w,x,y,z}  
\makeatletter
\newcommand\footnoteref[1]{\protected@xdef\@thefnmark{\ref{#1}}\@footnotemark}
\makeatother
\newcommand{\MyMapTemplateNoPrefix}[3]{\expandafter#1\csname#3\endcsname{#2{#3}}}
\forcsvlist{\MyMapTemplatePrefixc {\def} {\mathbf}{mb}} {0,1,a,b,c,d,e,f,g,h,i,j,k,l,m,n,o,p,q,r,s,u,v,w,x,y,z}  
\forcsvlist{\MyMapTemplatePrefixc {\def} {\mathbf}{mb}} {A,B,C,D,E,F,G,H,I,J,K,L,M,N,O,P,Q,R,S,T,U,V,W,X,Y,Z}

\DeclareMathOperator{\drop}{DropOut}
\DeclareMathOperator{\fdrop}{F-Drop}
\DeclareMathOperator{\fnoise}{F-Noise}
\DeclareMathOperator{\vat}{I-VAT}
\DeclareMathOperator{\objmask}{Obj-Msk}
\DeclareMathOperator{\conmask}{Con-Msk}
\DeclareMathOperator{\cutout}{G-Cutout}
\DeclareMathOperator{\confmask}{Conf-Mask}
\DeclareMathOperator{\abce}{ab-CE}
\DeclareMathOperator{\pairwise}{P-Wise}

\DeclareMathOperator{\power}{power}
\DeclareMathOperator{\maxiter}{max\_iter}
\DeclareMathOperator{\iter}{iter}

\begin{document}

\title{Semi-Supervised Semantic Segmentation with Cross-Consistency Training}

\author{
  Yassine Ouali\hspace{3pc}
  Céline Hudelot\hspace{3pc}
  Myriam Tami\\
  Université Paris-Saclay, CentraleSupélec, MICS, 91190, Gif-sur-Yvette, France.\\
  {\tt\small \{yassine.ouali,celine.hudelot,myriam.tami\}@centralesupelec.fr}
}

\maketitle

\begin{abstract}
	In this paper, we present a novel cross-consistency based semi-supervised
	approach for semantic segmentation. 
	Consistency training has proven to be a powerful semi-supervised learning 
	framework for leveraging unlabeled data under the cluster assumption,
	in which the decision boundary should lie in low density regions.
	In this work, we first observe that for semantic segmentation,
	the low density regions are more apparent
	within the hidden representations than within the inputs.
	We thus propose cross-consistency training, where an invariance of the predictions
	is enforced over different perturbations applied to the outputs of the encoder.
	Concretely, a shared encoder and a main decoder are
	trained in a supervised manner using the available labeled examples.
	To leverage the unlabeled examples, we enforce a consistency between 
	the main decoder predictions and those of the auxiliary decoders,
	taking as inputs different perturbed versions of the encoder's output, 
	and consequently, improving the encoder's representations.
	The proposed method is simple and
	can easily be extended to use additional
	training signal, such as image-level labels or pixel-level
	labels across different domains.
	We perform an ablation study to tease apart the effectiveness
	of each component, and conduct
	extensive experiments to demonstrate that
	our method achieves state-of-the-art results in several
	datasets.
	\footnote{Code available at: \url{https://github.com/yassouali/CCT}}
\end{abstract}
\vspace{-15pt}


\section{Introduction}
\label{intro}

In recent years, with the wide adoption of deep supervised learning 
within the computer vision community, significant strides were made 
across various visual tasks yielding impressive results. However, training 
deep learning models requires
a large amount of labeled data which acquisition is often costly and time consuming.
In semantic segmentation, given how expensive and laborious
the acquisition of pixel-level labels is,
with a cost that is 15 times and 60 times larger than that of region-level
and image-level labels respectively \cite{COCO}, the need
for data efficient semantic segmentation methods
is even more evident.

As a result, a growing attention is drown on deep
Semi-Supervised learning (SSL) to take advantage of a large amount of
unlabeled data and limit the need for labeled examples.
The current dominant SSL methods in deep learning
are consistency training
\cite{ladder_nets,TEnsembling,MeanTeachers,VAT}, pseudo labeling \cite{pseudo_labling},
entropy minimization \cite{entropy_mini} and bootstrapping
\cite{qiao2018deep}. Some newly introduced
techniques are based on generative modeling \cite{kumar2017semi,souly2017semi}.

\begin{figure}[t]
  \centering
  \includegraphics[width=\linewidth]{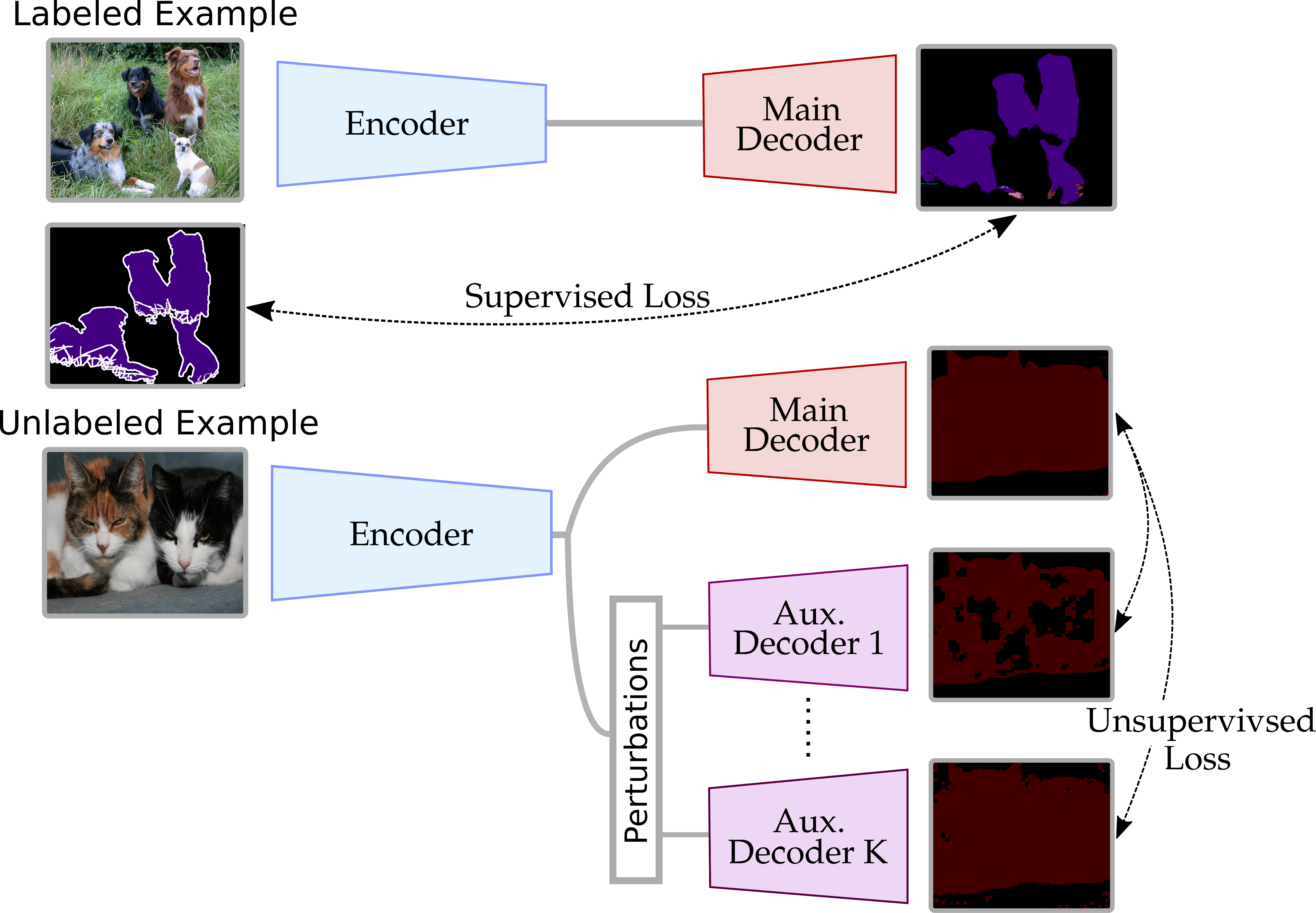}
  \caption{\textbf{The proposed Cross-Consistency training (CCT).} For the labeled examples,
  the encoder and the main decoder are trained in a supervised manner.
  For the unlabeled examples, a consistency between the main decoder's
  predictions and those of the auxiliary decoders is enforced, over
  different types of perturbations applied to the inputs
  of the auxiliary decoders. Best viewed in color.}
  \label{fig:overview}
  \vspace{-0.15in}
\end{figure}

However, the recent progress in SSL was confined to classification tasks,
and its application in semantic segmentation is
still limited. Dominant approaches \cite{DSRG,dilated_cam,erasing,lee2019ficklenet}
focus on weakly-supervised learning which principle is to generate
pseudo pixel-level labels by leveraging the weak labels, 
that can then be used, together with the limited strongly labeled examples, to train a segmentation
network in a supervised manner.
Generative Adversarial Networks (GANs) were
also adapted for SSL setting \cite{souly2017semi,hung2018adversarial}
by extending the generic GAN framework
to pixel-level predictions. The discriminator is then jointly
trained with an adversarial loss and a supervised loss over the labeled examples.


Nevertheless, these approaches suffer from some limitations.
Weakly-supervised approaches require weakly labeled examples along with
pixel-level labels, hence, they do not exploit the unlabeled data to 
extract additional training signal.
Methods based on adversarial training exploit the unlabeled
data, but can be harder to train.


To address these limitations, we propose
a simple consistency based semi-supervised method for semantic segmentation.
The objective in consistency training is to enforce an invariance 
of the model's predictions over small perturbations
applied to the inputs. 
As a result,
the learned model will be robust to such small changes.
The effectiveness of consistency training depends heavily on the 
behavior of the data distribution, \ie, the cluster assumption, where
the classes must be separated by low density regions. In semantic segmentation,
we do not observe the presence of low density regions separating
the classes within the inputs,
but rather within the encoder's outputs.
Based on this observation, we propose to enforce the consistency
over different forms of perturbations applied to the encoder's output.
Specifically,
we consider a shared encoder and a main decoder that are trained using 
the labeled examples. To leverage unlabeled data,
we then consider multiple auxiliary decoders whose inputs are perturbed
versions of the output of the shared encoder.
The consistency is imposed between the main decoder's predictions
and that of the auxiliary decoders (see \cref{fig:overview}).
This way, the shared encoder's representation is enhanced by 
using the additional training signal extracted from the unlabeled
data.
The added auxiliary decoders have a negligible amount 
of parameters compared to the encoder. Additionally, during inference,
only the main decoder is used, reducing the computation overhead
both in training and inference.


The proposed method is simple and efficient, it is also
flexible since it can easily be extended to use additional weak
labels and pixel-level labels across different domains
in a semi-supervised domain adaption setting.
With extensive experiments, we demonstrate the effectiveness of our approach on
PASCAL VOC \cite{pascalvoc} in a semi-supervised setting, and CityScapes,
CamVid \cite{camvid} and SUN \cite{sun} in a semi-supervised 
domain adaption setting. We obtain competitive results 
across different datasets and training settings.


\vspace{0.05in}
Concretely, our contributions are four-fold:
\begin{itemize}
\itemsep0em 
	\item We propose a cross-consistency training (CCT) method for semi-supervised semantic segmentation,
	where the invariance of the predictions is enforced over different perturbations injected into
	the encoder's output.
	\item We propose and conduct an exhaustive study of various types of perturbations.
	\item We extend our approach to use weakly-labeled data,
	and exploit pixel-level labels across different domains
	to jointly train the segmentation network.
	\item We demonstrate the effectiveness of our approach with
	with an extensive and detailed experimental results,
	including a comparison with the state-of-the-art, as well as
	an in-depth analysis of our approach with a detailed ablation study.
\end{itemize}


\section{Related Work}
\label{relatedwork}

\textbf{Semi-Supervised Learning.} 
Recently, many efforts have been made to adapt classic
SSL methods to deep learning, such as pseudo labeling \cite{pseudo_labling},
entropy minimization \cite{entropy_mini} and
graph based methods \cite{liu2018deep,kipf2016semi} in order to overcome this weakness.
In this work, we focus mainly on consistency training. We refer the reader to
\cite{chapelle2009semi} for a detailed overview of the field.
Consistency training methods are based
on the assumption that, if a realistic form of perturbation was applied to the unlabeled examples,
the predictions should not change significantly.
Favoring models with decision boundaries that reside in low density regions,
giving consistent predictions
for similar inputs. For example, $\Pi$-Model \cite{TEnsembling}
enforces a consistency over two perturbed versions of
the inputs under different data augmentations and dropout.
A weighted moving average of either the previous predictions
(\ie, Temporal Ensembling \cite{TEnsembling}),
or the model's parameters 
(\ie, Mean Teacher \cite{MeanTeachers}),
can be used
to obtain more stable predictions over the unlabeled examples.
Instead of relying on random perturbations, Virtual Adversarial Training (VAT) \cite{VAT}
approximates the perturbations that will alter the model's predictions the most.

Similarly, the proposed method enforces a consistency of predictions between the main decoder
and the auxiliary decoders over different perturbations, that are applied to the 
encoder's outputs rather than the inputs.
Our work is also loosely related to Multi-View learning \cite{multi-view-leaning} and
Cross-View training \cite{clark2018semi},
where each input to the auxiliary decoders
can be view as an alternate, but corrupt representation of the unlabeled examples.


\textbf{Semi-Supervised Semantic Segmentation.} 
A significant number of approaches use a limited number pixel-level labels
together with a larger number of inexact annotations,
\eg, region-level \cite{song2019boxdriven,Dai_2015} or
image-level labels \cite{lee2019ficklenet,CAM,dilated_cam,wheretolook}.
For image-level based weak-supervision, primary localization maps are
generated using class activation mapping (CAM) \cite{CAM}.
After refining the generated maps, they can
then be used to train a segmentation network together with
the available pixel-level labels in a SSL setting.

Generative modeling can also be used 
for semi-supervised semantic segmentation \cite{souly2017semi,hung2018adversarial}
to take advantage of the unlabeled examples.
Under a GAN framework,
the discriminator's predictions are extended over pixel classes,
and can then be jointly trained
with a Cross-Entropy loss over the labeled examples and an adversarial loss over the 
whole dataset.

In comparison, the proposed method exploits the unlabeled examples by enforcing
a consistency over multiple perturbations on the hidden representations level.
Enhancing the encoder's representation and the overall performance,
with a small additional cost in terms of computation and memory requirements.   

Recently, CowMix \cite{french2019consistency}, a concurrent method 
was introduced. CowMix, using MixUp \cite{zhang2017mixup}, enforces a consistency between the mixed outputs and the prediction over the mixed inputs. 
In this context, CCT differs as follows:
(1) CowMix, as traditional consistency regularization methods,
applies the perturbations to the inputs, but uses MixUp as a high-dimensional
perturbation to overcome the absence of the cluster assumption.
(2) Requires multiple forward passes though the network for one training iteration.
(3) Adapting CowMix to other settings (\eg, over multiple domains, using weak labels)
may require significant changes.
CCT is efficient and can easily be extended to other settings.


\textbf{Domain Adaptation.} 
In many real world cases, the existing discrepancy between
the distribution of training data and 
and that of testing data will often hinder the performances.
Domain adaptation aims to rectify this mismatch and tune 
the models for a better generalization at test time \cite{patel2015visual}.
Various generative and discriminative domain adaptation methods  
have been proposed for classification 
\cite{geng2011daml,ganin2014unsupervised,ganin2016domain,cao2018partial}
 and semantic segmentation 
 \cite{hoffman2016fcns,zhang2017curriculum,saleh2018effective,kalluri2018universal} tasks.

In this work, we show that enforcing a consistency 
across different domains can push the model toward better generalization,
even in the extreme case of non-overlapping label spaces.


\section{Method}
\label{method}


\subsection{The cluster assumption in semantic segmentation}
\label{clusterasump}
We start with our observation and analysis of the cluster
assumption in semantic segmentation, motivating the proposal of our cross-consistency training
approach. 
A simple way to 
examine it is to estimate
the local smoothness by measuring the local variations between
the value of each pixel and its local neighbors.
To this end, we compute the 
average euclidean distance at each spatial location and its 8 intermediate neighbors,
for both the inputs and the hidden representations
(\ie, the ResNet's \cite{resnet} outputs of a DeepLab v3 
\cite{deeplabv3} trained on COCO \cite{COCO}). For the inputs,
following \cite{french2019consistency}, we compute the average distance of a patch centered
at a given spatial location and its neighbors to simulate a realistic receptive 
field. For the hidden representations, we first upsample the feature map
to the input size, and then compute the average distance
between the neighboring activations ($2048$-dimensional feature vectors).
The results are illustrated in \cref{fig:clusterassup}.
We observe that the cluster assumption is violated at the input level, 
given that the low density regions do not align with the class boundaries.
On the contrary, for the encoder's outputs, 
the cluster assumption is maintained where the class
boundaries have high average distance,
thus corresponding to low density regions.
This observation motivates the following approach, in which
the perturbations are applied to the encoder's outputs rather than 
the inputs.


\subsection{Cross-Consistency Training for semantic segmentation}
\label{cctmathod}

\subsubsection{Problem Definition}
\label{prbldef}
In SSL, we are provided with a small set of labeled training
examples and a larger set of unlabeled training examples. Let 
$\cD_l=\{(\mbx_{1}^l, y_{1}), \ldots, (\mbx_{n}^l, y_{n})\}$ represent
the $n$ labeled examples and  $\cD_{u}=\{\mbx_1^u, \ldots, \mbx_m^u\}$
represent the $m$ unlabeled examples,
with $\mbx_i^u$ as the $i$-th unlabeled input image, and
$\mbx_i^l$ as the $i$-th labeled input image with spatial dimensions $H \times W$
and its corresponding pixel-level label $y_i \in \mathbb{R}^{C \times H \times W}$,
where $C$ is the number of classes.

\begin{figure}[t]
  \centering
  \includegraphics[width=\linewidth]{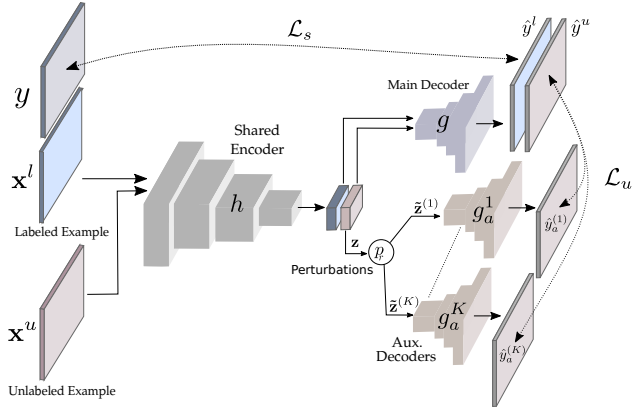}
  \caption{\textbf{Illustration of our approach.} For one training iteration, we sample
  a labeled input image $\mbx^l$ and its pixel-level label $y$ together with an unlabeled
  image $\mbx^u$. We pass both images through the encoder and main decoder,
  obtaining two main predictions $\hat{y}^l$ and $\hat{y}^u$.
  We compute the supervised loss using the pixel-level label $y$ and $\hat{y}^l$.
  We apply various perturbations to $\mbz$, the output of the encoder for $\mbx^u$,
  and generate auxiliary predictions $\hat{y}^{(i)}_a$
  using the perturbed versions $\tilde{\mbz}^{(i)}$. The unsupervised
  loss is then computed between the outputs of the auxiliary decoders and
  that of the main decoder.}
  \label{fig:CCT}
  \vspace{-0.15in}
\end{figure}

As discussed in the introduction, the objective is to exploit 
the larger number of unlabeled examples ($m \gg n$) to
train a segmentation network $f$,
to perform well on the test data drawn
from the same distribution as the training data. 
In this work, our architecture (see \cref{fig:CCT}) is composed of a shared encoder $h$
and a main decoder $g$, which constitute the segmentation network $f = g \circ h$.
We also introduce a set of $K$ auxiliary decoders $g_{a}^k$, with $k \in [1, K]$. 
While the segmentation network $f$ is trained on the labeled set $\cD_l$
in a traditional supervised manner,
the auxiliary networks $g_{a}^k \circ h$ are
trained on the unlabeled set $\cD_{u}$ 
by enforcing a consistency of predictions 
between the main decoder and the auxiliary decoders.
Each auxiliary decoder takes as input a perturbed version of the encoder's output,
and the main encoder is fed the uncorrupted intermediate representation.
This way, the representation learning of the encoder $h$ is further enhanced
using the unlabeled examples, and subsequently, that of
the segmentation network $f$. 
\vspace{-0.08in}

\begin{figure}
  \centering
  \includegraphics[width=\linewidth]{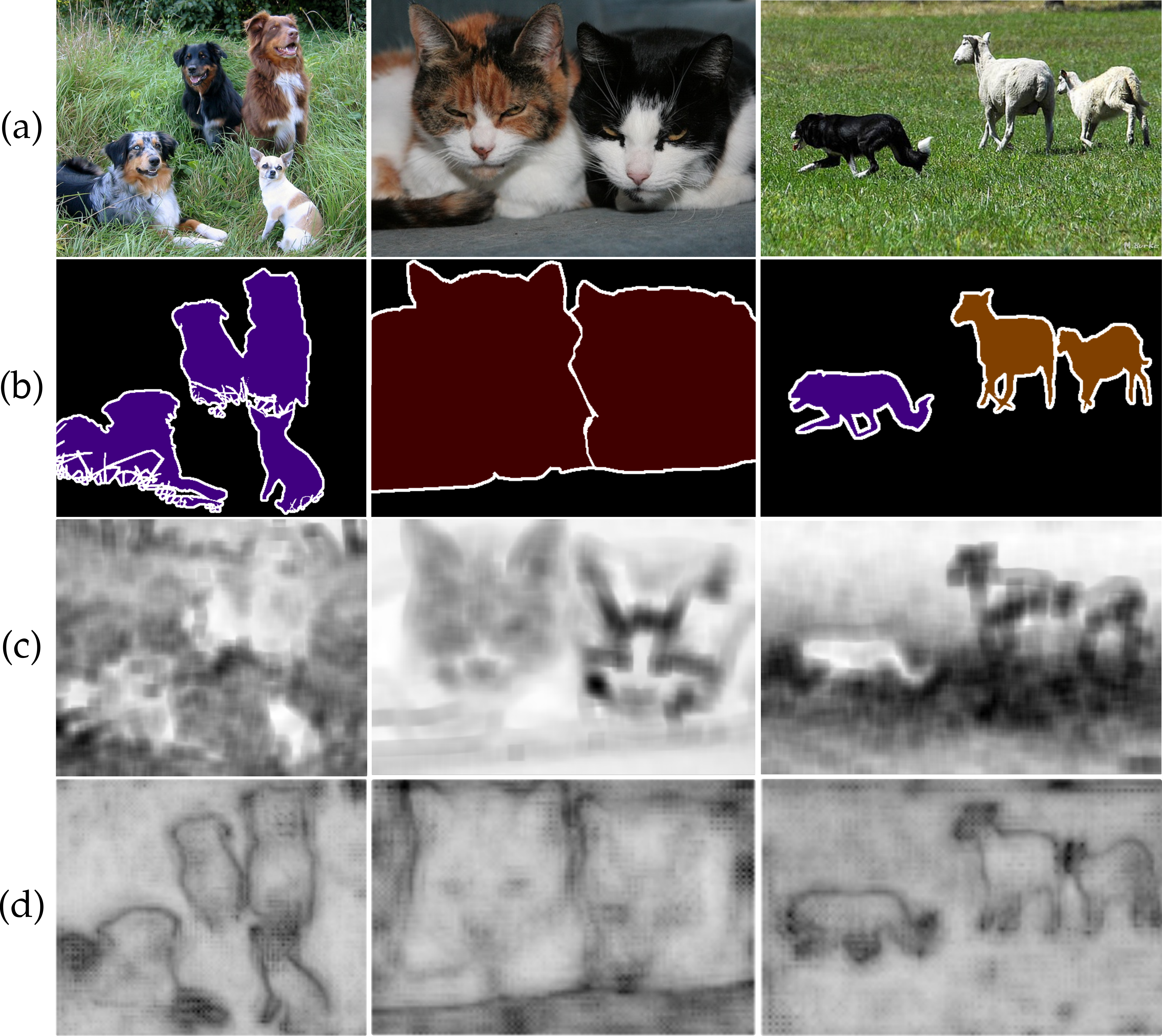}
  \caption{\textbf{The cluster assumption in semantic segmentation.}
  (a) Examples from PASCAL VOC 2012 \textit{train set}. (b) Pixel-level
  labels. (c) \textit{Input level.} The average euclidean distance
  between each patch of size $20 \times 20$
  centered at a given spatial location extracted from the input images,
  and its 8 neighboring patches. (d) \textit{Hidden representations level.} The
  average euclidean distance between a
  given $2048$-dimensional activation at each spatial
  location and its 8 neighbors. Darkest regions
  indicate high average distance.}
  \label{fig:clusterassup}
  \vspace{-0.15in}
\end{figure}

\subsubsection{Cross-Consistency Training}
\label{CCT}

As stated above, to extract additional training signal from the unlabeled set $\cD_u$, 
we rely on enforcing a consistency between the outputs of the main decoder $g_m$
and those of auxiliary decoders $g_{a}^k$.
Formally, for a labeled training example $\mbx_i^l$, and its pixel-level label $y_i$,
the segmentation network $f$ is trained using a Cross-Entropy (CE) based supervised loss 
$\cL_s$:
\begin{equation}
\cL_s=\frac{1}{|\cD_l|} \sum_{\mbx_i^l, y_i
\in \cD_{l}} \mbH(y_i, f(\mbx_i^l))
\label{eq:1}
\end{equation}
with $\mbH(.,.)$ as the CE.
For an unlabeled example $\mbx_i^u$, an intermediate
representation of the input is computed using the shared encoder $\mbz_i = h(\mbx_i^u)$\footnote{
Throughout the paper, $\mbz$ always refers to the output of the encoder
corresponding to an unlabeled input image $\mbx^u$.}.
Let us consider $R$ stochastic perturbations functions,
denoted as $p_r$ with $r \in [1, R]$,
where one perturbation function can be assigned to multiple auxiliary decoders.
With various perturbation settings, we generate $K$ perturbed versions $\tilde{\mbz}_i^k$
of the intermediate representation $\mbz_i$,
so that the $k$-th perturbed version is to be fed
to the $k$-th auxiliary decoder.
For consistency, we consider the perturbation function as
part of the auxiliary decoder,
(\ie, $g_{a}^k$ can be seen as $g_{a}^k \circ p_r$).
The training objective is then to minimize the unsupervised loss $\cL_u$,
which measures the discrepancy
between the main decoder's output and that of the auxiliary decoders:
\begin{equation}
\cL_u=\frac{1}{|\cD_u|} \frac{1}{K} \sum_{\mbx_i^u \in \cD_{u}}
\sum_{k=1}^{K} \mbd(g(\mbz_i), g_{a}^k(\mbz_i))
\label{eq:2}
\end{equation}
with $\mbd(.,.)$ as a distance measure between
two output probability distributions (\ie, the outputs of a $softmax$
function applied over the channel dimension).
In this work, we choose to use mean squared error (MSE) as a distance measure.

The combined loss $\cL$ for consistency based SSL is then computed as:
\begin{equation}
\cL = \cL_{\text{s}} + \omega_u \cL_{\text{u}}
\label{eq:3}
\end{equation}
where $\omega_u$ is an unsupervised loss weighting function.
Following \cite{TEnsembling},
to avoid using the initial noisy predictions of the main encoder,
$\omega_u$ ramps up
starting from zero along a Gaussian curve up to a fixed weight $\lambda_u$.
Concretely, at each training
iteration, an equal number of examples are sampled from the
labeled $\cD_l$ and unlabeled $\cD_u$ sets. The supervised loss is computed using 
the main encoder's output and pixel-level labels. For the unlabeled examples,
we compute the MSE between the prediction of each auxiliary decoder and that 
of the main decoder. The total loss is then compute and back-propagated to train 
the segmentation network $f$ and the auxiliary networks $g_a^k \circ h$.
Note that the unsupervised loss $\cL_u$ is not back-propagated
through the main-decoder $g$, only the labeled examples are used to train $g$.

\subsubsection{Perturbation functions}
\label{perturbations}
An important factor in consistency training is the perturbations 
to apply to the hidden representation, \ie., the encoder's output $\mbz$.
We propose three types of perturbation functions $p_r$:
feature based, prediction based and random. 

\vspace{0.05in}
\textbf{Feature based perturbations.}
They consist of either injecting noise into or dropping some of
the activations of encoder's output feature map $\mbz$.

\begin{itemize}
\vspace{-0.12in}
\itemsep0em 
	\item $\fnoise$:
	we uniformly sample a noise tensor $\mbN \sim \mathcal{U}(-0.3, 0.3)$
	of the same size as $\mbz$.
	After adjusting its amplitude by multiplying it with $\mbz$,
	the noise is then injected into the
	encoder's output $\mbz$ to get $\tilde{\mbz} = (\mbz \odot \mbN) + \mbz$. This way,
	the injected noise is proportional to each activation. 

	\item $\fdrop$: we first uniformly sample
	a threshold $\gamma \sim \mathcal{U}(0.6, 0.9)$. After summing over
	the channel dimension and normalizing the 
	feature map $\mbz$ to get $\mbz'$, we generate a mask
	$\mbM_{\text{drop}} = \{\mbz' < \gamma\}_{\mathbf{1}}$\footnote{$
	\{\text{condition}\}_{\mathbf{1}}$ is a boolean function outputting
	1 if the condition is true, 0 otherwise.},
	which is then used to obtain the perturbed version
	$\tilde{\mbz} = \mbz \odot \mbM_{\text{drop}}$.
	This way, we mask
	$10\%$ to $40\%$ of the most active regions in the feature map.
\end{itemize}

\textbf{Prediction based perturbations.}
They consist of adding perturbations based
on the main decoder's prediction $\hat{y} = g(\mbz)$
or that of the auxiliary decoders.
We consider masking based perturbations
($\conmask$, $\objmask$ and $\cutout$)
in addition to adversarial perturbations ($\vat$).

\begin{itemize}
\itemsep0em 
	\item Guided Masking: Given the importance of context relationships for
	complex scene understanding \cite{oliva2007role},
	the network might be too reliant on these relationships. 
	To limit them, we create two perturbed versions of 
	$\mbz$ by masking the detected objects ($\objmask$) and the context ($\conmask$).
	Using $\hat{y}$, we generate an object mask $\mbM_{\text{obj}}$
	to mask the detected foreground objects
	and a context mask $\mbM_{\text{con}} = 1 - \mbM_{\text{obj}}$, which are
	then applied to $\mbz$.

	\item Guided Cutout ($\cutout$): 
	in order to reduce the reliance on 
	specific parts of the objects, and inspired by Cutout \cite{cutout}
	that randomly masks some parts of the input image,
	we first find the possible spatial
	extent (\ie., bounding box) of each detected object using $\hat{y}$.
	We then zero-out a random crop within each object's bounding box
	from the corresponding feature map $\mbz$.

	\item Intermediate VAT ($\vat$):
	to further push the output distribution to be isotropically smooth around each
	data point, we investigate using VAT \cite{VAT} as a perturbation function
	to be applied to $\mbz$ instead of the unlabeled inputs.
	For a given auxiliary decoder, we find the adversarial perturbation $r_{adv}$
	that will alter its prediction the most. The noise is then injected into
	$\mbz$ to obtain the perturbed version $\tilde{\mbz} = r_{adv} + \mbz$.
\end{itemize}

\textbf{Random perturbations.} ($\drop$) Spatial dropout  \cite{spatial_dropout} is also applied to $\mbz$ as a random perturbation.


\subsubsection{Practical considerations}
A each training iteration, we sample an equal number of labeled
and unlabeled samples. As a consequence,
we iterate on the set $\cD_l$ more times than on 
its unlabeled counterpart $\cD_u$,
thus risking an overfitting of the labeled set $\cD_l$.

\textbf{Avoiding Overfitting}.
Motivated by \cite{pohlen2017full} who observed improved results by
sampling only $6\%$ of the hardest pixels, and \cite{xie2019unsupervised} who
showed an improvement when gradually releasing the supervised training signal in a SSL setting,
we propose an annealed version of the bootstrapped-CE ($\abce$) in
\cite{pohlen2017full}.
With an output $f(\mbx_i^l) \in \mathbb{R}^{C \times H \times W}$
in the form of a probability distribution over the pixels, we only
compute the supervised loss over the pixels
with a probability  less than a threshold $\eta$:
\begin{equation}
\cL_s=\frac{1}{|\cD_l|} \sum_{\mbx_i^l, y_i 
\in \cD_{l}} \{f(\mbx_i^l) < \eta\}_{\mathbf{1}} \mbH(y_i, f(\mbx_i^l))
\label{eq:4}
\end{equation}

To release the supervised training signal,
the threshold parameter $\eta$ is gradually 
increased from $\frac{1}{C}$
to $0.9$ during the beginning of training,
with $C$ as the number of output classes.


\subsection{Exploiting weak-labels}
\label{WSL}
In some cases, we might be provided with additional training data
that is less expensive to acquire compared to pixel-level labels,
\eg, image-level labels.
Formally, instead of an unlabeled set $\cD_u$,
we are provided with a weakly labeled set
$\cD_w=\{(\mbx_{1}^w, y_{1}^w), \ldots, (\mbx_{m}^w, y_{m}^w)\}$
alongside a pixel-level labeled set $\cD_l$, with $y_{i}^w$
is the $i$-th image-level label corresponding to the $i$-th
weakly labeled input image $\mbx_{i}^w$.
The objective is to
extract additional information from the weak labeled set $\cD_w$
to further enhance the representations of the encoder $h$.
To this end, we add a classification branch $g_c$
consisting of a global average pooling layer followed by
a classification layer,
and pretrain the encoder for a classification task
using binary CE loss.



Following previous works \cite{ahn2018learning,lee2019ficklenet,DSRG},
the pretrained encoder and the added
classification branch can then be exploited to
generate pseudo pixel-level labels $y_p$.
We start by generating
the CAMs $M$ as in \cite{CAM}.
Using $M \in \mathbb{R}^{C \times H \times W}$, 
we can then generate pseudo labels $y_p$,
with a background $\theta_{bg}$ and a foreground $\theta_{fg}$
thresholds. The pixels
with attention scores less than $\theta_{bg}$ (\eg, $0.05$)
are considered as background.
For the pixels with an attention score larger than $\theta_{fg}$ 
(\eg, $0.30$), they are assigned the
class with the maximal attention score,
and the rest of the pixels are ignored.
After generating $y_p$, we conduct a final refinement
step using dense CRF \cite{krahenbuhl2011efficient}.

In addition to considering $\cD_w$ as an unlabeled set
and imposing a consistency over its examples,
the pseudo-labels are used to train the auxiliary networks $g_a^k \circ h$
using a weakly supervised loss $\cL_w$.
In this case, the loss in \cref{eq:3} becomes:
\begin{equation}
\cL = \cL_{\text{s}} + \omega_u \cL_u + \omega_w \cL_w
\end{equation}
With
\begin{equation}
\cL_w=\frac{1}{|\cD_w|} \frac{1}{K} \sum_{\mbx_i^w \in \cD_{w}}
\sum_{k=1}^{K} \mbH(y_p, g_{a}^k(\mbz_i))
\end{equation}


\subsection{Cross-Consistency Training on Multiple Domains}
\label{DACT}

In this section, we extend the propose framework to a 
semi-supervised domain adaption setting. We consider the 
case of two datasets $\{\cD^{(1)}, \cD^{(2)}\}$ with partially
or fully non-overlapping label spaces, each
one contains a set of labeled and unlabeled examples 
$\cD^{(i)} = \{\cD^{(i)}_l, \cD^{(i)}_u\}$. The objective 
is to simultaneously train a segmentation network to do well on the test
data of both datasets, which is drown from the different distributions.

\begin{figure}[t]
  \centering
  \includegraphics[width=\linewidth]{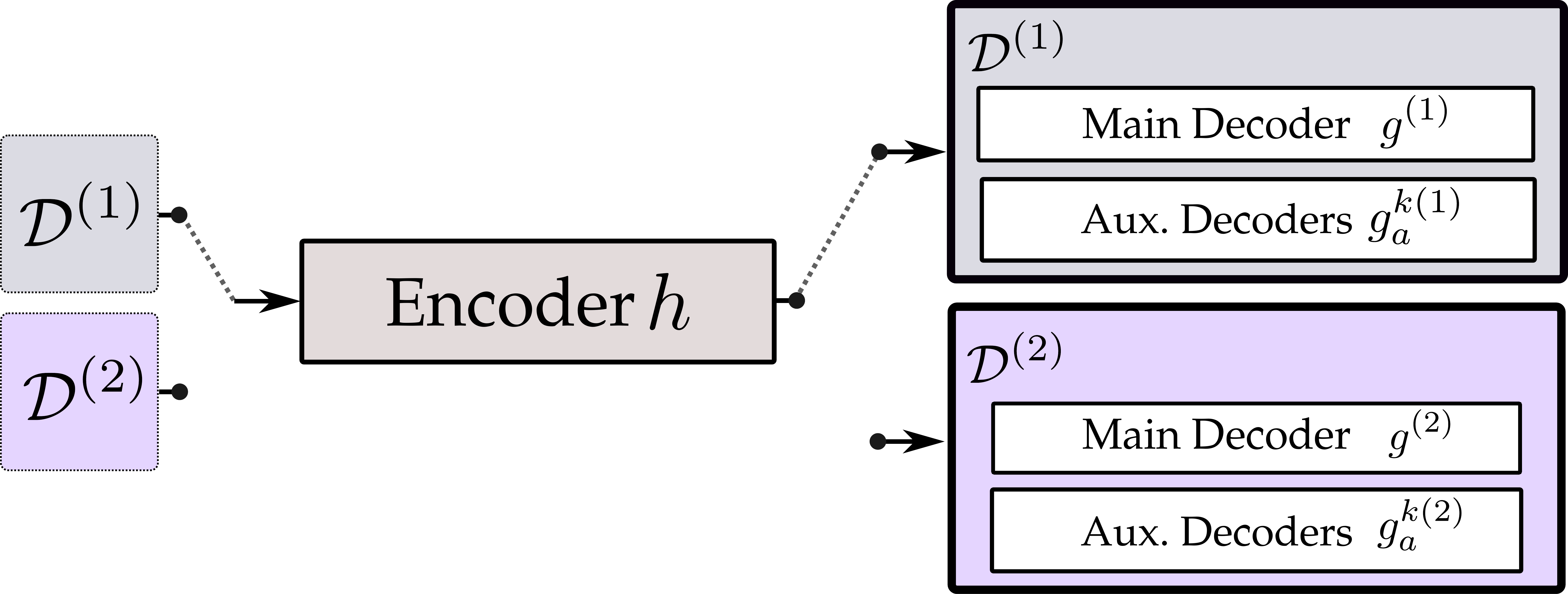}
  \caption{\textbf{CCT on multiple domains}. On top of a shared
  encoder, we add domain specific main decoder and $K$ auxiliary decoders.
  During training, we alternate between the two domains,
  sampling labeled and unlabeled examples and training the corresponding
  decoders and the shared encoder at each iteration.}
  \label{fig:da}
  \vspace{-0.15in}
\end{figure}

Our assumption is that enforcing 
a consistency over both unlabeled sets $\cD^{(1)}_u$ and $\cD^{(2)}_u$
might impose an invariance of the encoder's representations across the 
two domains. To this end, on top of the shared encoder $h$, we add 
domain specific main decoder $g^{(i)}$ and auxiliary decoders $g_a^{k(i)}$.
Specifically, as illustrated in \cref{fig:da},
we add two main decoders and $2K$ auxiliary decoders on top of
the encoder $h$. During
training, we alternate between the two datasets, at each iteration, sampling an equal
number of labeled and unlabeled examples from each one,
computing the loss in \cref{eq:3} and training
the shared encoder and the corresponding main and 
auxiliary decoders.

\section{Experiments}
\label{exp}

\begin{figure*}[t]
  \centering
  \includegraphics[width=\linewidth]{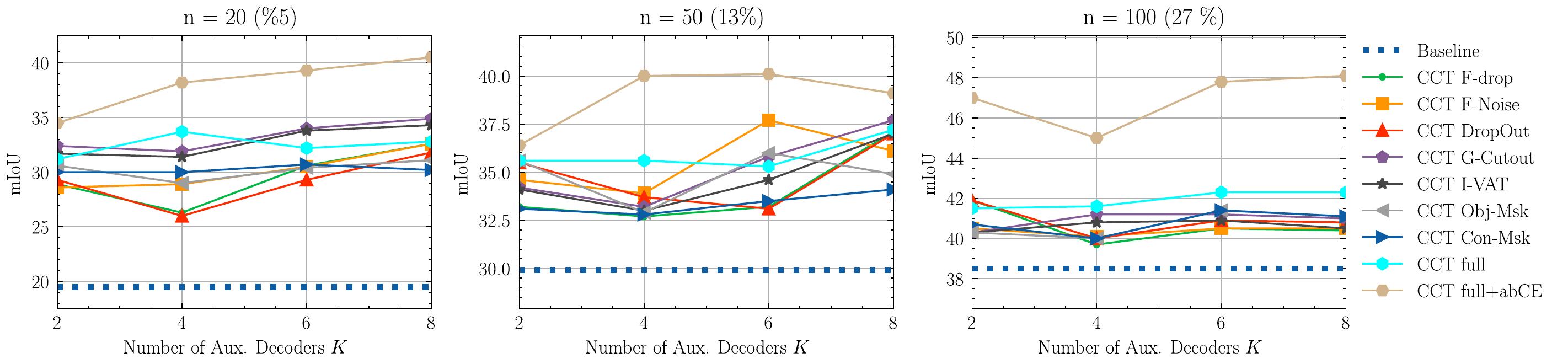}
  \caption{\textbf{Ablation Studies on CamVid with 20, 50 and 100 labeled images}.
  With different types of perturbations and a variable number
  of auxiliary decoders $K$, we compare the individual and the combined
  effectiveness of the perturbations to the baseline in which the model
  is trained only on the labeled examples. CCT full refers to using all of the 7 perturbations, \ie. 
  the number of auxiliary decoder is $K \times 7$.}
  \label{fig:ablationcam}
  \vspace{-0.15in}
\end{figure*}

\begin{figure}[t]
  \centering
  \includegraphics[width=\linewidth]{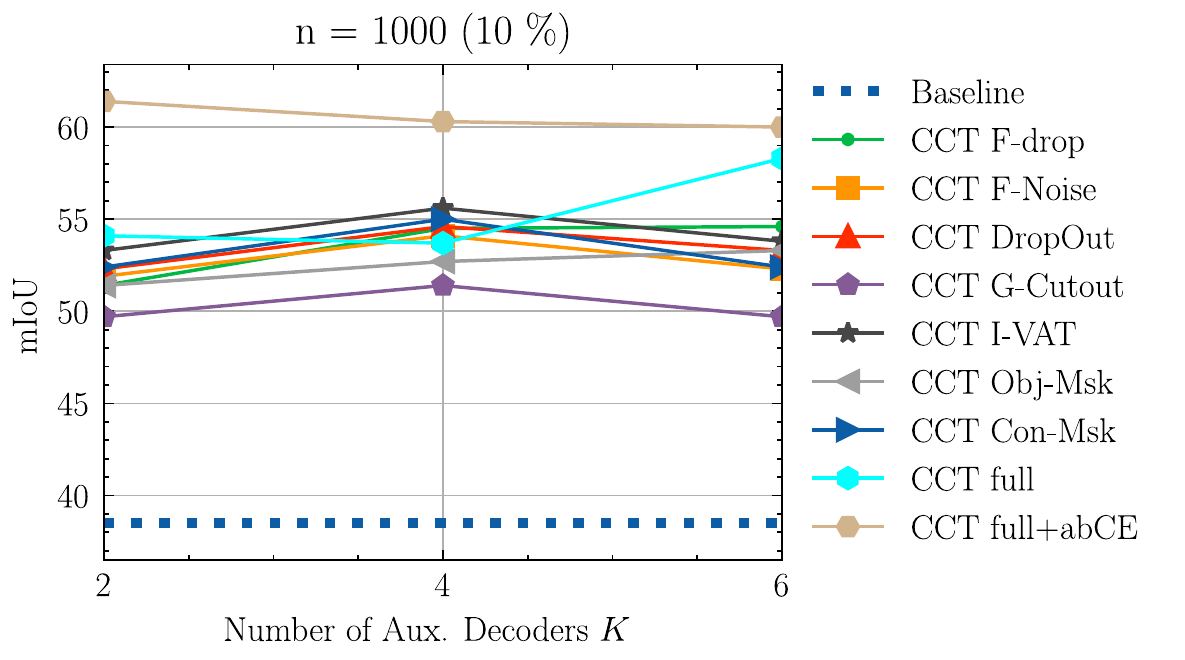}
  \caption{\textbf{Ablation study on PASCAL VOC}. Ablation study results with
  1000 labeled examples using different perturbations and various numbers of 
  auxiliary decoders $K$.}
  \label{fig:ablationvoc}
  \vspace{-0.15in}
\end{figure}

To evaluate the proposed method and investigate its effectiveness in different settings,
we carry out detailed experiments. In \cref{sec:ssl-exp}, we present an extensive
ablation study to highlight the contribution of each component within the proposed framework,
and compare it to state-of-the-art methods in a semi-supervised setting. Additionally,
in \cref{sec:da-exp} we apply the proposed method in a
semi-supervised domain adaptation setting
and show performance above baseline methods.


\subsection{Network Architecture}
\textbf{Encoder.}
For the following experiments,
the encoder is based on a ResNet-50 \cite{resnet}
pretrained on ImageNet \cite{imagenet} provided by \cite{you2019torchcv}
and a PSP module \cite{PSPnet}. Following previous works
\cite{PSPnet,DSRG,ahn2018learning},
the last two strided convolutions of ResNet are replaced with dilated convolutions.

\textbf{Decoders.} For the decoders, taking the efficiency and the number of parameters 
into consideration, we choose to only use $1\times1$ convolutions.
After an initial $1\times1$ convolution to adapt the depth
to the number of classes $C$,
we apply a series of three sub-pixel convolutions \cite{shi2016real}
with ReLU non-linearities to upsample the outputs
to original input size. 


\subsection{Datasets and Evaluation Metrics}
\textbf{Datasets.} In a semi-supervised setting,
we evaluate the proposed method on PASCAL VOC \cite{pascalvoc}, consisting of 
21 classes (with the background included) and
three splits, training, validation and testing,
with of $1464$, $1449$ and $1456$ images respectively.
Following the common practice \cite{DSRG,PSPnet}, we augment
the training set with additional images from \cite{hariharan2011semantic}.
Note that the pixel-level labels are only extracted from the original training set.

For semi-supervised domain adaption, for partially overlapping label spaces,
we train on both Cityscapes \cite{cityscapes} and CamVid \cite{camvid}.
Cityscapes is a finely annotated autonomous driving dataset with $19$
classes. We are provided with three splits,
training, validation and testing with $ 2975$, $500$ and $1525$ images respectively.
CamVid contains 367 training, 101 validation and 233 testing images.
Although originally the dataset is labeled with $38$ classes, we use
the $11$ classes version \cite{segnet}. For experiments over
non-overlapping labels spaces, we train on Cityscapes and SUN RGB-D \cite{sun}.
SUN RGB-D is an indoor segmentation dataset with $38$ classes
containing two splits, training and validation, with
$5285$ and $5050$ images respectively.
Similar to \cite{kalluri2018universal}, we train on
the $13$ classes version \cite{handa2016understanding}.

\textbf{Evaluation Metrics.} 
We report the results using mIoU (\ie, mean of class-wise intersection over union)
for all the datasets.

\subsection{Implementation Details}

\textbf{Training Settings.}
The implementation is based on the PyTorch 1.1 \cite{pytorch} framework.
For optimization, we train for $50$ epochs using
SGD with a learning rate of $0.01$ and
a momentum of $0.9$. During training, the learning rate is annealed following the \textit{poly}
learning rate policy, where at each iteration, the base learning rate is multiplied by
$1-(\frac{\iter}{\maxiter})^\power$ with $\power = 0.9$.

For PASCAL VOC,
we take crops of size $321 \times 321$ and apply random rescaling in the 
range of $[0.5, 2.0]$ and random horizontal flip.
For Cityscapes, Cam-Vid and SUN RGB-D, following
\cite{kalluri2018universal,hung2018adversarial},
we resize the input images to $512 \times 1024$,
$360 \times 480$ and $480 \times 640$ respectively,
without any data-augmentation.


\textbf{Reproducibility} All the experiments are conducted on a V-100 GPUs.
The implementation is available at: \url{https://github.com/yassouali/CCT}

\textbf{Inference Settings.}
For PASCAL VOC, Cityscapes and SUN RGB-D, we 
report the results obtained on the validation set,
and on the test set of CamVid dataset.

\subsection{Semi-Supervised Setting}
\label{sec:ssl-exp}
\subsubsection{Ablation Studies}

The proposed method consists of several types of perturbations
and a variable number of auxiliary decoders. We thus start by studying the 
effect of the perturbation functions with
different numbers of auxiliary decoders, in order to provide additional
insight into their individual performance and their combined 
effectiveness. Specifically, we measure the effect of different numbers
of auxiliary decoders $K$ (\ie, $K=2$, $4$, $6$ and $8$) of a given perturbation type.
We refer to this setting of our method as “CCT \{\textit{perturbation type}\}”,
with seven possible perturbations.
We also measure the combined effect of all perturbations resulting
in $K \times 7$ auxiliary decoders in total, and refer to it as “CCT full”.
Additionally, “CCT full+$\abce$”
indicates the usage of the annealed-bootstrapped CE as a supervised loss function.
We compare them to the baseline, in which the model is trained 
only using the labeled examples.


\begin{table}[!t]
	\centering
	\resizebox{0.45\textwidth}{!}{
		\begin{tabular}{l p{2cm} p{2cm} c}
		\toprule \\[-1em]
		Method & Pixel-level Labeled Examples & Image-level Labeled Examples & Val \\
		\midrule \\[-1.5ex]
		WSSL~\cite{papandreou2015weakly} & 1.5k & 9k &  64.6 \\ 
		GAIN~\cite{wheretolook} & 1.5k & 9k &  60.5 \\ 
		MDC~\cite{dilated_cam} & 1.5k & 9k &  65.7 \\
		DSRG~\cite{DSRG} & 1.5k & 9k &  64.3 \\ 
		Souly \etal~\cite{souly2017semi} & 1.5k & 9k &  65.8 \\
		FickleNet~\cite{lee2019ficklenet} & 1.5k & 9k &  65.8 \\
		\midrule
		Souly \etal~\cite{souly2017semi} & 1.5k & - &  64.1 \\
		Hung \etal~\cite{hung2018adversarial} & 1.5k & - &  68.4 \\
		\midrule
		CCT & 1k & - &  64.0 \\ 
		CCT & 1.5k & - &  69.4 \\ 
		CCT & 1.5k & 9k & \textbf{73.2} \\ 
		\bottomrule \\[-1.5ex]
		\end{tabular}}
	\vspace{0.05in}
	\caption{\textbf{Comparison with the-state-of-the-art.}
	CCT performance on PASCAL VOC compared to other semi-supervised 
	approaches.}
	\label{tab:pascal}
	\vspace{-0.1in}
\end{table}

\textbf{CamVid.} We carried out the ablation on CamVid with 20, 50 and 100 labels;
the results are shown in \cref{fig:ablationcam}.
We find that each perturbation outperforms the baseline,
with the most dramatic differences in the 20-label setting with up to $21$ points.
We also surprisingly observe an insignificant overall
performance gap among different perturbations,
confirming the effectiveness of enforcing
the consistency over the hidden representations for semantic segmentation,
and highlighting the versatility of CCT and its success
with numerous perturbations.
Increasing $K$ results in a modest improvement overall, with the smallest change
for $\conmask$ and $\objmask$ due to their lack of stochasticity.
Interestingly, we also observe
a slight improvement when combining all of the perturbations, 
indicating that the encoder is able to generate representations
that are consistent over many perturbations, and subsequently,
improving the overall performance.
Additionally, 
gradually releasing the training signal using $\abce$
helps increase the performance with up to $8\%$,
which confirms that overfitting of the labeled examples can
cause a significant drop in performance.

\textbf{PASCAL VOC.} 
In order to investigate
the success of CCT on larger datasets,
we conduct additional ablation experiments 
on PASCAL VOC using 1000 labeled examples, 
The results are summarized in \cref{fig:ablationvoc}.
We see similar results, where the proposed method
makes further improvement compared to the baseline with
different perturbations, from $10$ to $15$ points.
The combined perturbations yield a small increase
in the performance, with the biggest difference
with $K = 6$.
Furthermore, similar to CamVid, when using the $\abce$ loss,
we see a significant gain
with up to $7$ points compared to CCT full.

Based on the conducted ablation studies, for the rest of the experiments, we use
the setting of “CCT full” 
with $K=2$ for $\conmask$ and $\objmask$ due to their lack of stochasticity,
$K=2$ for $\vat$ given its high computational cost, and $K=6$ for 
the rest of the perturbations, and refer to it as “CCT”.

\vspace{-0.1in}
\subsubsection{Comparison to Previous Work}
To further explore the effectiveness of our framework,
we quantitatively compare it with previous semi-supervised
semantic segmentation methods on PASCAL VOC.
\cref{tab:pascal} compares CCT with other semi-supervised approaches.
Our approach outperforms previous works relying on the same level
of supervision and even methods which exploit image-level labels. 
We also observe an increase of $3.8$ points when using additional 
image-level labels, affirming the flexibility of CCT, and
the possibility of using it with different types of labels
without any learning conflicts.

\subsection{Semi-Supervised Domain Adaptation Setting}
\label{sec:da-exp}
In real world applications, we are often provided with pixel-level labels
collected from various sources, thus distinct data distributions. 
To examine the effectiveness of CCT when applied 
to multiple domains with a variable degree of labels overlap,
we train our model simultaneously on two datasets,
Cityscapes (CS) + CamVid (CVD) for partially overlapping labels, and  
Cityscapes + SUN RGB-D (SUN) for the disjoint case.

\begin{table}[H]
\centering
\resizebox{0.45\textwidth}{!}{
	\begin{tabular}{@{} l c c c c c c c @{}} 
		\toprule  \\[-1em]
		\multirow{2}{*}{Method} &
		\multicolumn{3}{c}{n=50}  && \multicolumn{3}{c}{n=100} \\
		\cmidrule{2-4} \cmidrule{6-8}
		& CS &  CVD & Avg. && CS & CVD & Avg. \\
		\midrule \\[-1.5ex]
		Kalluri, \etal~\cite{kalluri2018universal} & 34.0 & 53.2 & 43.6 && 41.0 & 54.6 & 47.8 \\ 
		\midrule
		Baseline & 31.2 & 40.0 & 35.6 && 37.3 & 34.4 & 35.9 \\ 
		CCT & 35.0 & 53.7 & \textbf{44.4} && 40.1 & 55.7 & \textbf{47.9} \\ 
		\bottomrule \\[-1.5ex]
	\end{tabular}}
	\vspace{0.05in}
		\caption{\textbf{CCT applied to CS+CVD.}
		CCT performance when simultaneously trained on two datasets
		with overlapping label spaces, which are Cityscapes (CS)
		and CamVid (CVD).}
		\label{tab:cscam}
	\vspace{-0.1in}
\end{table}

\textbf{Cityscapes + CamVid.}
The results for CCT on Cityscapes and CamVid datasets with
50 and 100 labeled examples are given in \cref{tab:cscam}. 
Similar to the SSL setting, CCT outperforms the baseline significantly, where
the model is iteratively trained using only on the labeled examples,
with up to 12 points for $n=100$, we even see a modest increase compared
to previous work. This confirms our hypothesis
that enforcing a consistency over different
datasets does indeed push the encoder to produce invariant representation
across different domains, and consequently, increases the performance over the baseline
while delivering similar results on each domain individually.

\begin{table}[H]
	\centering
	\resizebox{0.45\textwidth}{!}{
		\begin{tabular}{l p{2cm} c c c c }
		\toprule  \\[-1em]
		Method & Labeled Examples && CS & SUN & Avg.   \\
		\midrule \\[-1.5ex]
		SceneNet~\cite{mccormac2017scenenet} & Full (5.3k) && - & 49.8 & - \\
		Kalluri, \etal~\cite{kalluri2018universal} & 1.5k && 58.0 & 31.5 & 44.8 \\
		\midrule
		Baseline & 1.5k && 54.3 & 38.1 & 46.2   \\ 
		CCT & 1.5k && 58.8 &  45.5 & \textbf{52.1} \\ 
		\bottomrule \\[-1.5ex]

		\end{tabular}}
	\vspace{0.05in}
	\caption{\textbf{CCT applied to CS+CVD.} CCT performance when
	trained on both datasets Cityscapes (CS)
		and SUN RGB-D (SUN) datasets, for the case
		of non-overlapping label spaces.}
	\label{tab:cssun}
	\vspace{-0.1in}
\end{table}

\textbf{Cityscapes + SUN RGB-D.}
For cross domain experiments, where the two domains have
distinct labels spaces, we train on both Cityscapes and SUN RGB-D
to demonstrate the capability of CCT to extract
useful visual relationships and perform knowledge transfer between
dissimilar domains, even in completely different settings. The results are
shown in \cref{tab:cscam}. Interestingly, despite the distribution mismatch
between the datasets, and the high number of labeled examples ($n=1500$),
CCT still provides a meaningful boost over the baseline with 5.9 points difference
and 7.3 points compared to previous work.
Showing that, by enforcing a consistency of predictions on the unlabeled sets
of the two datasets over different
perturbations, we can extract additional
training signal and enhance the representation learning of the encoder,
even in the extreme case with non-overlapping label spaces, without any
performance drop when an invariance of representations across both datasets
is enforced at the level of encoder's outputs.



\section{Conclusion}
In this work, we present cross-consistency training (CCT),
a simple, efficient and flexible method for
a consistency based semi-supervised
semantic segmentation, yielding state-of-the-art results.
For future works, a possible direction is exploring the usage of other perturbations
to be applied at different levels within the segmentation network.
It would also be interesting to adapt and
examine the effectiveness of CCT in other visual tasks and learning settings,
such as unsupervised domain adaptation.


\vspace{0.1in}

{\footnotesize
\textbf{Acknowledgements.}
{This work was supported by Randstad corporate research chair. We would also like to thank
Saclay-IA plateform of Université Paris-Saclay and Mésocentre computing center of CentraleSupélec and
École Normale Supérieure Paris-Saclay for providing the computational resources.}
}



{\small
\bibliographystyle{ieee_fullname}
\bibliography{paper}
}


\setcounter{section}{0}
\renewcommand\thesection{\Alph{section}}
\renewcommand\thesubsection{\thesection.\arabic{subsection}}

\vspace{0.3in}
\begin{center}
  \Large\textbf{Supplementary Material}\\
\end{center}

\section{Comparison with Traditional Consistency Training Methods}
In this section, we present the experiments to validate the
observation that for semantic segmentation,
enforcing a consistency over different perturbations applied to the 
encoder's outputs rather than the inputs is more aligned with
the cluster assumption. To this end, we compare
the proposed method with traditional consistency based SSL methods.
Specifically, we conduct experiments using VAT \cite{VAT} and Mean Teachers
\cite{MeanTeachers}. In VAT, at each training iteration, the unsupervised
loss is computed as the KL-divergence between the model's predictions
of the input $\mbx^u$ and its perturbed version $\mbx^u + r_{adv}$. For
Mean Teachers, the discrepancy is
measured using Mean Squared Error (MSE) between the prediction of
the model and the prediction using an exponential weighted version of it.
In this case, the
noise is sampled at each training step with SGD.

\begin{table}[H]
	\centering
	\resizebox{0.45\textwidth}{!}{
		\begin{tabular}{p{3cm} p{3cm} c}
		\toprule \\[-1em]
		Splits & n=500 & n=1000 \\
		\midrule
		Baseline & 51.4 & 59.2  \\
		\midrule
		Mean Teachers & 51.3 & 59.4 \\ 
		VAT & 50.0 & 57.9 \\
		CCT & \textbf{58.6} & \textbf{64.4} \\
		\bottomrule \\[-1.5ex]
		\end{tabular}}
	\vspace{0.05in}
	\caption{\textbf{CCT compared to traditional consistency methods.}
	We conduct an ablation study on PASCAL VOC,
	where we compare the performance of the baseline
	to the proposed method CCT, VAT and Mean Teachers.
	$n$ represents the number of labeled examples.}
	\label{tab:consbaselines}
	\vspace{-0.1in}
\end{table}

The results are presented  in \cref{tab:consbaselines}.
We see that applying the adversarial noise
to inputs with VAT results in lower performance compared to the baseline.
When using Mean Teachers, in which the noise is not implicitly added
to the inputs, we obtain similar performance to the baseline.
These results confirm our observation that enforcing a consistency
over perturbations applied to the hidden representations
is more aligned with the cluster assumption,
thus yielding better results.

\section{Additional Results and Evaluations}
\subsection{Distance Measures}
In the experiments presented in \cref{sec:ssl-exp}, MSE was used
as a distance measure $\mbd(.,.)$ for the unsupervised loss $\cL_u$,
to measure the discrepancy between the main and auxiliary predictions.
In this section, we investigate the 
effectiveness of other 
distance measures between the output probability distributions. Specifically,
we compare the performance of MSE to the KL-divergence and the JS-divergence.
For an unlabeled example $\mbx^u$, we obtain a main prediction $y^u$ with 
the main decoder and an auxiliary prediction $y_a^k$
with a given auxiliary decoder $g_a^k$. We compare the following distance measures:
\begin{equation}
\mbd_{\text{MSE}}(y^u,y_a^k) = \frac{1}{N} \sum_{i}^N (y^u(i) - y_a^k(i))^2
\end{equation}
\begin{equation}
\mbd_{\text{KL}}(y^u,y_a^k) =  \frac{1}{N} \sum_{i}^N y^u(i) \log \frac{y^u(i)}{y_a^k(i)}
\end{equation}
\begin{equation}
\mbd_{\text{JS}}(y^u,y_a^k) = \frac{1}{2} \mbd_{\text{KL}}(y^u,m) + \frac{1}{2} \mbd_{\text{KL}}(y_a^k,m) 
\end{equation}
where $m = \frac{1}{2} (y^u(i)+y_a^k(i))$ and $y^{*}(i)$ refers to the output
probability distribution
at a given spatial location $i$. The results of the comparison are shown
in \cref{tab:distmeasures}.

\begin{table}[H]
	\centering
	\resizebox{0.45\textwidth}{!}{
		\begin{tabular}{p{3cm} p{3cm} c}
		\toprule \\[-1em]
		Splits & n=500 & n=1000 \\
		\midrule
		Baseline & 51.4 & 59.2  \\
		\midrule
		CCT \footnotesize{KL} & 54.0 & 62.5 \\
		CCT \footnotesize{JS} & 58.4 & 64.3 \\
		CCT \footnotesize{MSE} & \textbf{58.6} & \textbf{64.4} \\
		\bottomrule \\[-1.5ex]
		\end{tabular}}
	\vspace{0.05in}
	\caption{\textbf{CCT with different distance measures.}
	We compare the performance of MSE to the KL-divergence and
	the JS-divergence on PASCAL VOC dataset.}
	\label{tab:distmeasures}
	\vspace{-0.1in}
\end{table}

We observe similar performance with
$\mbd_{\text{MSE}}$ and $\mbd_{\text{JS}}$,
while we only obtain 2.6 and 3.3 points gain for $n=500$
and $n=1000$ respectively 
over the baseline when using $\mbd_{\text{KL}}$.
The low performance of $\mbd_{\text{KL}}$ might be due to its non-symmetric nature.
With $\mbd_{\text{KL}}$, the auxiliary decoders are heavily penalized over sharp
but wrong predictions, thus pushing them to produce uniform and uncertain outputs,
and reducing the amount of training signal that can be extracted from the unlabeled examples.
However, with $\mbd_{\text{JS}}$, which is a symmetrized and smoothed version
of $\mbd_{\text{KL}}$, we can bypass the zero avoidance nature of the KL-divergence.
Similarly, $\mbd_{\text{MSE}}$ can be seen as a multi-class Brier score \cite{MixMatch} which is
less sensitive to completely incorrect predictions,
giving it similar properties to $\mbd_{\text{JS}}$ with a lower computational cost.

\subsection{Confidence Masking and Pairwise Loss}
\paragraph{Confidence Masking. ($\confmask$)} 
When training on the unlabeled examples, we use the main predictions as
the source for consistency training, which may result in a corrupted training
signal when based on uncertain predictions.
A possible way to avoid this is masking the uncertain
predictions. Given a main prediction $y^u$ in the form of a probability
distribution over the classes $C$
at different spatial locations $i$. We
compute the unsupervised loss $\cL_u$ only 
over the pixels $i$ with probability $\max\limits_C y^u(i)$ greater than
a fixed threshold $\beta$ (\eg, 0.5).

\paragraph{Pairwise Loss.  ($\pairwise$)} 
In CCT, we enforce the consistency of predictions only between
the main and auxiliary decoders, without
any pairwise consistency in between the auxiliary predictions.
To investigate the effectiveness of
enforcing such an additional pairwise consistency,
we add the following an additional loss term $\cL_{\pairwise}$
to the total loss in \cref{eq:3}
to penalize the auxiliary predictive variance:
\begin{equation}
\cL_{\pairwise} = \frac{1}{K} \sum_{k=1}^K (y_a^k - \bar{y}_a)^2
\label{eq:pw}
\end{equation}
with $\bar{y}_a$ as the mean of the auxiliary predictions $y_a^k$.
Given $K$ auxiliary decoders, the computation of $\cL_{\pairwise}$
is in the order of $K^2$. To reduce it, at each
training iteration, we only compute $\cL_{\pairwise}$ over a 
randomly chosen subset of
the auxiliary predictions (\eg, $8$ out of $K=30$).

\cref{tab:confmask} shows the results of the experiments when using CCT
with $\confmask$ and $\pairwise$. Interestingly, we do not
observe any gain over CCT when using $\confmask$, indicating that using
the uncertain main predictions to enforce the consistency
does not hinder the performance. Additionally, adding a pairwise loss term
results in lower performance compared to CCT, with $3$ and $3.2$ points difference 
in both settings, indicating that adding $\cL_{\pairwise}$ can potentially
compel the auxiliary decoders to produce similar predictions regardless of the
applied perturbation, thus diminishing the representation learning 
of the encoder, and the performance of the segmentation network as a whole.

\begin{table}[H]
	\centering
	\resizebox{0.45\textwidth}{!}{
		\begin{tabular}{p{4cm} p{2cm} l}
		\toprule \\[-1em]
		Splits & n=500 & n=1000 \\
		\midrule
		Baseline & 51.4 & 59.2  \\
		\midrule
		CCT \footnotesize{+$\confmask$} & 58.4 & 63.3\\
		CCT \footnotesize{+$\cL_{\pairwise}$} & 55.6 & 61.2 \\
		CCT & \textbf{58.6} & \textbf{64.4} \\
		\bottomrule \\[-1.5ex]
		\end{tabular}}
	\vspace{0.05in}
	\caption{\textbf{CCT with $\pairwise$ and $\confmask$}.
	The results of the effect of adding a confidence masking over unsupervised
	loss and a pairwise loss between the auxiliary predictions on PASCAL VOC \textit{val} set.}
	\label{tab:confmask}
\end{table}


\section{Algorithm}
The proposed Cross-Consistency training method can be summarized by the following
Algorithm:

\removelatexerror
\begin{algorithm}[H]
{\SetAlgoNoLine
\SetKwInput{kwRequire}{Require}
\KwIn{Labeled image $\mbx^l$, its pixel-level label $y$ and unlabeled image $\mbx^u$}
\kwRequire{Shared encoder $h$, main decoder $g_m$, $K$ auxiliary decoders $g_a^k$}
\vspace{10pt}
1) Forward $\mbx^l$ through the encoder and main decoder: $\hat{y}^l = g_m(h(\mbx^l))$\\
2) Forward the unlabeled input through the shared encoder: $\mbz = h(\mbx^u)$\\
3) Generate the main decoder's prediction for $\mbx^u$: $\hat{y}^u = g_m(\mbz)$\\
4) Generate the aux. decoders predictions for $\mbx^u$:\\
\Indp	
\For{$k$ in $[1, K]$}{
- Apply a given perturbation $\tilde{\mbz} = p_l(z)$ \\
- Forward through the aux. decoder $k$: $\hat{y}_a^i = g_a^k(\tilde{\mbz})$ \\
}
\Indm
5) Training the network.\\
\Indp
$\cL_s = \mbH(\hat{y}^u, y)$\\
$\cL_u = \frac{1}{K} \sum_k \mbd(\hat{y}^u, \hat{y}_a^k)$\\
Update network by $\cL= \cL_s + \omega_u \cL_u$
 \caption{\label{algo}Cross-Consistency Training (CCT).}}
\end{algorithm}

\section{Further Investigation of The Cluster Assumption}
The learned feature of a
CNNs are generally more homogeneous, and at higher
layers, the network learns to compose low level features into semantically
meaningful representations while discarding high-frequency information (\eg, texture).
However, the leaned features in a segmentation network seem to have a unique property;
the class boundaries correspond to low density regions, which are not observed in networks
trained on other visual tasks (\eg, classification, object detection).
See \cref{fig:clusterassup} for an illustration of this difference.

\begin{figure*}
  \centering
  \includegraphics[width=0.9\linewidth]{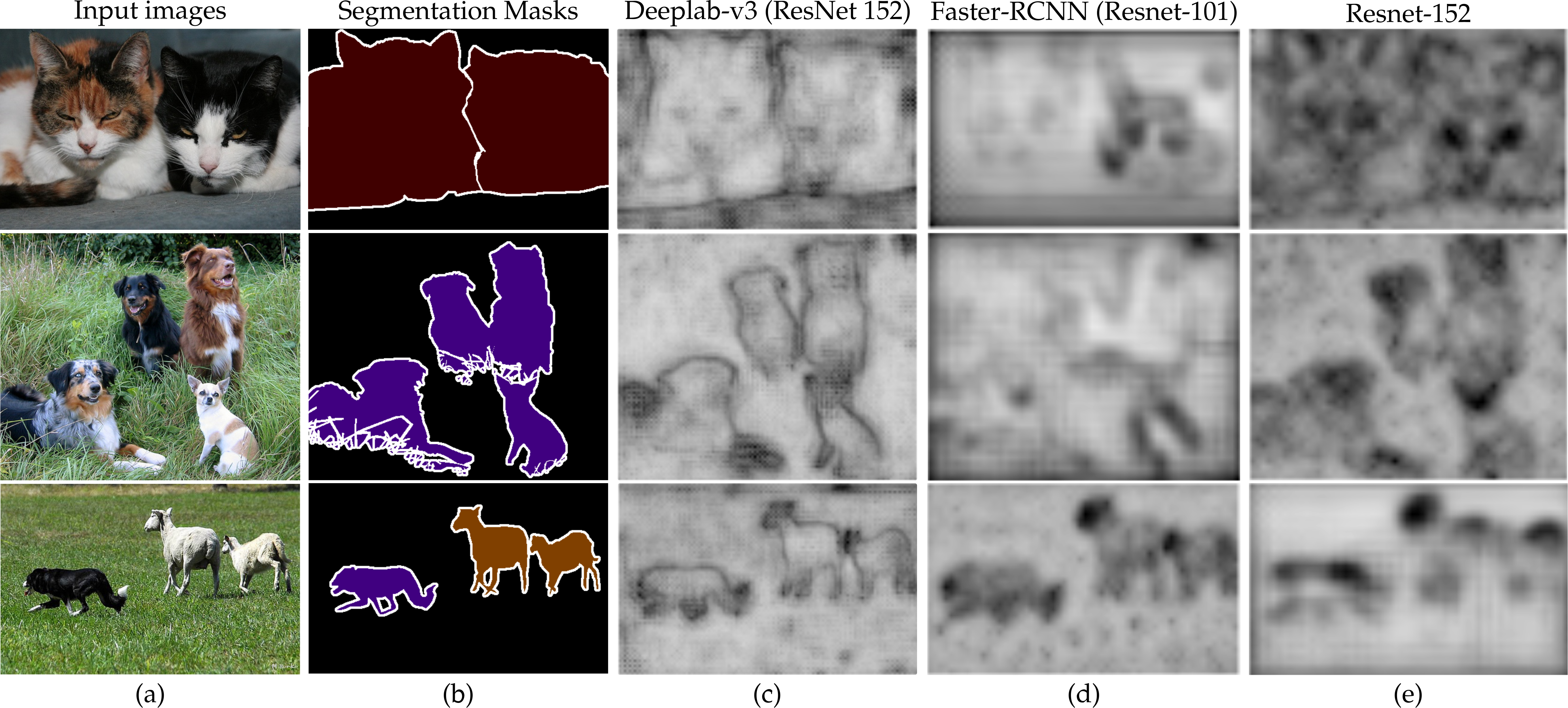}
  \caption{\textbf{The smoothness of CNNs features trained on different tasks.}
  (a) Examples from PASCAL VOC 2012 \textit{train set}. (b) Results for a segmentation network. 
  (c) Results for an object detection network. (b) Results for a classification network.}
  \label{fig:clusterassup}
  \vspace{-0.1in}
\end{figure*}

\section{Adversarial Distribution Alignment}
When applying CCT over multiple domains,
and to further reduce the discrepancy between the encoder's representations
of the two domains (\ie, the empirical distribution mismatch measured
by the $\cH$-Divergence \cite{ben2010theory}),
we investigate the addition of a discriminator branch $g_d$,
which takes as input the encoder's representation $\mbz$, and predict 0 for examples
from $\cD^{(1)}$ and 1 for examples from $\cD^{(2)}$. Hence, we add
the following adversarial loss to the total loss  in \cref{eq:1}:
\begin{dmath}
\cL_{adv} = \frac{1}{|\cD^{(1)}|} \sum_{\mbx_i \in \cD^{(1)}} \log(g_d(\mbz_i)) +
\frac{1}{|\cD^{(2)}|} \sum_{\mbx_i \in \cD^{(2)}} (1 - \log(g_d(\mbz_i)))
\label{eq:7}
\end{dmath}

The encoder and the discriminator branch are competitors within a 
min-max framework, \ie, the training objective is $\displaystyle \max_{g_d} \min_h \cL_{adv}$,
which can be directly optimized using a gradient reversal layer
as in \cite{ganin2014unsupervised}. The total loss in this case is:
\begin{equation}
\cL = \cL_s + \lambda_{adv} \cL_{adv} + \omega_u \cL_u 
\label{eq:8}
\end{equation}

\begin{table}[H]
\centering
\resizebox{0.45\textwidth}{!}{
	\begin{tabular}{@{} l c c c c c c c @{}} 
		\toprule  \\[-1em]
		\multirow{2}{*}{Method} &
		\multicolumn{3}{c}{n=50}  && \multicolumn{3}{c}{n=100} \\
		\cmidrule{2-4} \cmidrule{6-8}
		& CS &  CVD & Avg. && CS & CVD & Avg. \\
		\midrule \\[-1.5ex]
		Baseline & 31.2 & 40.0 & 35.6 && 37.3 & 34.4 & 35.9 \\ 
		CCT & 35.0 & 53.7 & \textbf{44.4} && 40.1 & 55.7 & \textbf{47.9} \\
		CCT \footnotesize{+$\cL_{avd}$} & 35.3 & 49.2 & 42.2 && 37.7 & 52.8 & 45.2 \\ 
		\bottomrule \\[-1.5ex]
	\end{tabular}}
	\vspace{0.05in}
		\caption{\textbf{CCT applied to CS+CVD.}}
		\label{tab:cscam}
	\vspace{-0.1in}
\end{table}

For the discriminator branch, similar to \cite{hung2018adversarial}, we use a fully 
convolutional discriminator, with a series of $3\times3$
convolutions and Leaky ReLU non-linearities as shown in \cref{tab:discriminator}.
The outputs are of the same size as the encoder outputs (\ie with an input image
of spatial dimensions $H \times W$, the outputs of $g_d$
are of size $2 \times \frac{H}{8} \times \frac{W}{8}$).

\begin{table}[H]
	\centering
	\resizebox{0.45\textwidth}{!}{
		\begin{tabular}{p{4cm} c}
		\toprule \\[-1em]
		Description & Resolution $\times$ channels \\
		\midrule
		Conv $3 \times 3 \times 64$ & $\tfrac{1}{8} \times 64$ \\
		LeakyReLU & \\
		Conv $3 \times 3 \times 128$ & $\tfrac{1}{8} \times 128$ \\
		LeakyReLU &  \\
		Conv $3 \times 3 \times 256$ & $\tfrac{1}{8} \times 256$ \\
		LeakyReLU &  \\
		Conv $3 \times 3 \times 512$ & $\tfrac{1}{8} \times 512$ \\
		LeakyReLU &  \\
		Conv $1 \times 1 \times 2$ & $\tfrac{1}{8} \times 2$ \\
		\bottomrule \\[-1.5ex]
		\end{tabular}}
	\caption{\textbf{Discriminator Branch.} The added discriminator branch
	on top of the encoder, in order to further push towards an invariance of the 
	encoder's representations between the different domains.}
	\label{tab:discriminator}
\end{table}

The results are shown in \cref{tab:discriminator}. Surprisingly, adding
a discriminator branch diminishes the performance of the segmentation network,
hinting to possible learning conflicts between CCT and the adversarial loss.

\section{Multi-scale Inference}
To further enhance the predictions of our
segmentation network, we conduct additional evaluations on PASCAL VOC using multi-scale
to simulate a similar situation to training where we apply random
scaling between $0.5$ and $2$, random croping and random horizontal flip.
We apply the same augmentations during test. For a given
test image, we create 5 versions using 5 scales: $0.5$,
$0.75$, $1$, $1.25$ and $1.5$, each image is also flipped horizontally,
resulting in $10$ versions of the test image. The model's prediction
are computed for each image, rescaled to the original size, and are
then aggregated by pixel-wise average pooling. The final result
is obtain by taking the $argmax$ over the classes for each spatial location.

In \cref{tab:multiscale}, we report the results obtained with multi-scale inference.
\begin{table}[H]
	\centering
	\resizebox{0.45\textwidth}{!}{
		\begin{tabular}{p{4cm} p{2cm} l}
		\toprule \\[-1em]
		 & n & mIoU \\
		\midrule
		CCT & 1000 									& 67.3 (+3.3) \\
		CCT & 1500 									& 73.4 (+4) \\
		CCT \footnotesize{+$9k$ Image-level labels} & 1500 & 75.1 (+2.9) \\
		\bottomrule \\[-1.5ex]
		\end{tabular}}
	\vspace{0.05in}
	\caption{\textbf{CCT results with multi-scale inference}. The mIoU
	when we apply multi-scale inference on PASCAL VOC \textit{val} set.}
	\label{tab:multiscale}
\end{table}

\section{Virtual Adversarial Training (VAT)}
Without the label information in a
semi-supervised setting, VAT \cite{VAT} lends itself as
a consistency regularization technique. It trains the
output distribution to be isotropically smooth around
each data point by selectively smoothing the model in
its most anisotropic direction. In our case, we apply
the adversarial perturbation $r_{adv}$ to the encoder
output $\mbz = h(\mbx^u)$.
For a given auxiliary decoder $g_a^k$,
we would like to compute the adversarial perturbation $r_{adv}$
that will alter its predictions the most.
We start by sampling a Gaussian noise $r$ of the same size as $\mbz$,
compute its gradients $grad_r$
with respect the loss between the two predictions, with and without the injections
of the noise $r$ (\ie, KL-divergence is used as a distance measure $\mbd(.,.)$).
$r_{adv}$ can then be obtained by normalizing and scaling $grad_r$ by a hyperparameter
$\epsilon$. This can be written as follows:
\begin{equation}
r \sim \mathcal{N}\left(0, \frac{\xi}{\sqrt{dim(\mbz)}} I \right)
\end{equation}
\begin{equation}
grad_r=\nabla_{r} \mbd\left(g_a^k(\mbz), g_a^k(\mbz+r)\right)
\end{equation}
\begin{equation}
r_{adv}=\epsilon \frac{grad_r}{\|grad_r\|}
\end{equation}

Finally, the perturbed input to $g_a^k$ is $\tilde{\mbz} = r_{adv} + \mbz$. 
The main drawback of such method is requiring multiple forward and backward passes
for each training iteration to compute $r_{adv}$. In our case, the
amount of computations needed are reduced
given the small size of the auxiliary decoders.

\section{Dataset sizes}
For the size of each split of the datasets used in our experiments, see \cref{tab:datasets}.
\begin{table}[H]	
	\centering
	\resizebox{0.45\textwidth}{!}{
		\begin{tabular}{l p{2cm} p{2cm} c}
		\toprule \\[-1em]
		Splits & Train & Val & Test \\
		\midrule
		PASCAL VOC & 10582 & 1449 &  1456 \\ 
		Cityscapes & 2975 & 500 & 1525 \\ 
		CamVid & 367 & 101 &  233 \\ 
		SUN RGB-D & 5285 & - &  5050 \\ 
		\bottomrule \\[-1.5ex]
		\end{tabular}}
	\vspace{0.05in}
	\caption{\textbf{Semantic Segmentation Datasets.}
	The size of each split of the datasets used in the experiments.}
	\label{tab:datasets}
\end{table}

\section{Further Experimental Details}
For the experiments throughout the paper, we used a ResNet 50 and a PSP module \cite{PSPnet}
for the encoder. As for the decoders, we used an initial $1 \times 1$ convolutions to adapt
the depth to the number of classes $C$, followed by a series of $1 \times 1$
sub-pixel convolutions \cite{shi2016real} (\ie, PixelShuffle) to upsample the feature maps
to the original size. For details see \cref{tab:architecture}.

\begin{table}[H]
\centering
\resizebox{0.45\textwidth}{!}{
	\begin{tabular}{@{} l c | l c @{}} 
		\toprule  \\[-1em]
		\multicolumn{2}{c}{Encoder} & \multicolumn{2}{c}{Decoder} \\
		\midrule
		Description & \shortstack{Resolution $\times$\\channels} & Description & \shortstack{Resolution $\times$\\channels} \\
		\midrule
		ResNet 50 & $\tfrac{1}{8} \times 2048$ &	 				Conv $1 \times 1 \times C$ & $\tfrac{1}{8} \times C$ \\
		PSPModule \cite{PSPnet} & $\tfrac{1}{8} \times 512$ &		Conv $1 \times 1 \times 4C$ & $\tfrac{1}{8} \times 4C$ \\
								&							&		PixelShuffle & $\tfrac{1}{4} \times C$ \\
								&							&		Conv $1 \times 1 \times 4C$ & $\tfrac{1}{4} \times 4C$ \\
								&							&		PixelShuffle & $\tfrac{1}{2} \times C$ \\
								&							&		Conv $1 \times 1 \times 4C$ & $\tfrac{1}{2} \times 4C$ \\
								&							&		PixelShuffle & $1 \times C$ \\

		\bottomrule \\[-1.5ex]
	\end{tabular}}
	\vspace{0.05in}
		\caption{\textbf{Encoder-Decoder architecture.} Showing the layer
		type, the number of the outputs channels and the spatial resolution.}
		\label{tab:architecture}
	\vspace{-0.1in}
\end{table}

\textbf{Inference Settings.}
For PASCAL VOC,
during the ablation studies reported in \cref{fig:ablationvoc},
in order to reduce the training time, we trained on smaller size image.
Specifically, we resize the bigger side to $300$
and randomly take crops of size $240 \times 240$.
For the comparisons with state-of-the-art
we resize the bigger side to $400$ and take crops of size 
$321 \times 321$ and conduct the inference on the original sized images.
For the rest of the datasets, the evaluation is conducted
on the same sizes as the ones used during training.

\section{Hyperparameters}
\label{sec:hyperparameters}
In order to present a realistic evaluation
of the proposed method, and following the practical considerations mentioned in \cite{oliver2018realistic}.
We avoid any form of intensive hyperparameter search,
be it that of the perturbation functions, model architecture or training settings.
We choose the hyperparameters that resulted in stable training by hand,
we do expect however that better performances can be achieved with
a comprehensive search.
The hyperparameters settings used in the experiments are summarized in \cref{tab:hyperparam}.

\begin{table}[H]
\centering
\small
\begin{tabular}{lc}
\toprule
\multicolumn{2}{c}{\textbf{Training}} \\
\midrule
\multicolumn{2}{c}{SGD} \\
        Learning rate & $10^{-2}$ \\
        Momentum & $0.9$ \\
        Weight Decay & $10^{-4}$ \\
\midrule
\multicolumn{2}{c}{Number of training epochs} \\
		PASCAL VOC & 50 \\
		CamVid & 50 \\
		Cityscapes \& CamVid & 50 \\
		Cityscapes \& SUN RGB-D & 100 \\
\midrule
\multicolumn{2}{c}{\textbf{Losses}} \\
\midrule
\multicolumn{2}{c}{Unsupervised loss $\cL_u$} \\
        Rampup periode for $\cL_u$ & 0.1 \\
        $\cL_u$ weight $\lambda_u$ & 30 \\
\midrule
\multicolumn{2}{c}{Weakly-supervised loss $\cL_w$} \\
        Rampup periode for $\cL_w$ & 0.1 \\
        $\cL_{w}$ weight $\lambda_w$ & 0.4 \\
\midrule
\multicolumn{2}{c}{Annealed Cross-Entropy loss $\abce$} \\
        Rampup periode & 0.5 \\
        Final threshold & 0.9 \\
\midrule
\multicolumn{2}{c}{Adversarial loss $\cL_{adv}$} \\
        Weight $\lambda_{adv}$ & $2.10^{-2}$ \\
\midrule
\multicolumn{2}{c}{\textbf{Perturbation Functions}} \\
\midrule
\multicolumn{2}{c}{\textbf{$\vat$}} \\
        VAT $\epsilon$ & 2.0 \\
        VAT $\xi$ & $10^{-6}$ \\
\midrule
\multicolumn{2}{c}{\textbf{$\drop$}} \\
        Dropout rate $p$ & $0.5$\\
\midrule
\multicolumn{2}{c}{\textbf{$\cutout$}} \\
        Area of the dropped region  & 0.4 \\
\midrule
\multicolumn{2}{c}{\textbf{$\fdrop$}} \\
        Drop threshold range & $[0.6, 0.9]$ \\
\midrule
\multicolumn{2}{c}{\textbf{$\fnoise$}} \\
        The uniform noise range & $[-0.3, 0.3]$ \\
\bottomrule
\end{tabular}
\vskip 0.1in
\caption{\textbf{Hyperparameters.} The hyperparameter
settings used in our experiments.}
\label{tab:hyperparam}
\end{table}


\section{Ramp-up functions}
\begin{figure}[H]
  \centering
  \includegraphics[width=\linewidth]{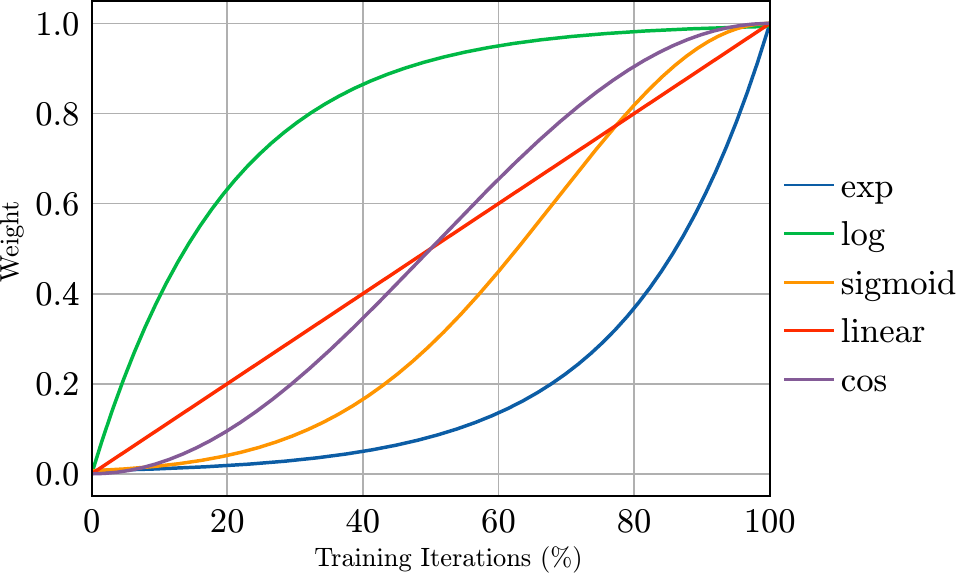}
  \caption{\textbf{Different ramp-up schedules.}}
  \label{fig:ramps}
  \vspace{-0.15in}
\end{figure}

For the unsupervised loss in \cref{eq:2}, the weighting function $w_u$
is gradually increased from $0$ up to a fixed final weight $\lambda_u$.
The rate of increase can follow many possible rates depending on
the schedule used. \cref{fig:ramps} shows different ramp-up schedules.
For our experiments, following \cite{TEnsembling},
$w_u$ ramps-up following an exp-schedule:
\begin{equation}
w_u(t) = \min(\lambda_u,  e^{5 (\frac{t}{T} - 1)} \times \lambda_u)
\end{equation}
with $t$ as the current training iteration and $T$ as the desired ramp-up length
(\eg, the first $10\%$ of training time).
Similarly, the threshold $\eta$ in the $\abce$ loss (\cref{eq:4}) is gradually increased
starting from $1/C$, with $C$ as the number of classes,
up to a final threshold $\alpha$ (\eg, $0.9$) within
a ramp-up period $T$ (\eg, the first $40\%$ of training time).
For $\abce$, we use a log-schedule to quickly
increase $\eta$ in the beginning of training:
\begin{equation}
\eta(t) = \min(\alpha, (1 - e^{-5 \frac{t}{T}}) \times (\alpha - 1/C) +1/C)
\end{equation}

\section{Computational Overhead}
\begin{table}[H]
\centering
\resizebox{0.45\textwidth}{!}{
	\begin{tabular}{l|c|l|l|l}
		\toprule  \\[-1em]
		Decoders         & Input size & GPU memory (MB) & GPU time (ms)\\
		\midrule
		Main Decoder 	& \multirow{8}{*}{$96 \times 96$} & 139 & 2.0 \\
		$\drop$			&    &  139  & 2.6  \\
		$\fdrop$		&    &  157  & 3.0  \\
		$\fnoise$		&    &  175  & 45.7  \\
		$\vat$			&    &  463  & 82.3  \\
		$\objmask$		&    &  463  & 2.7  \\	
		$\conmask$		&    &  463  & 2.4  \\	
		$\cutout$		&    &  463  & 3.3  \\	
		\midrule
		Main Decoder 	& \multirow{8}{*}{$256\times 128$} & 457 & 4.0 \\
		$\drop$			&    &  520  & 4.7  \\
		$\fdrop$		&    &  520  & 5.2  \\
		$\fnoise$		&    &  584  & 149.5  \\
		$\vat$			&    &  1592  & 176.0  \\
		$\objmask$		&    &  1592  & 4.7  \\	
		$\conmask$		&    &  1592  & 4.6  \\	
		$\cutout$		&    &  1592  & 7.1  \\	
		\bottomrule
	\end{tabular}}
	\vspace{0.1in}
	\caption{\textbf{Computation and memory statistics.}
	Comparisons between the main 
	and auxiliary decoders with different perturbation functions.
	The channel numbers of the input feature maps $\mbz$ is $512$.
	The lower the values, the better.
	\label{tab:memstats}}
	\vspace{-0.1in}
\end{table}

In order to present a comparison between the computational overhead of 
the different types of auxiliary decoders,
we present various computation and memory statistics in \cref{tab:memstats}.
We observe that for the majority of the auxiliary decoders, the GPU time is similar
to that of the main decoder. However, we see a significant increase for $\vat$ given
the multiple forward and backward passes required to compute the adversarial perturbation.
$\fnoise$ also results in high GPU time due to the sampling procedure. To this end
we reduce the number of $\vat$ decoders (\eg, $K=2$ for our experiments).
For $\fnoise$, for an input tensor of size
$B \times C \times H \times W$ with $B$ as the batch size,
instead of sampling a noise tensor $\mbN$ of the same
size, we sample a tensor of size $1 \times C \times H \times W$
and apply it over the whole batch.
Significantly reduces the computation the computation time without impacting the performance.


\section{Change Log}
\noindent \textbf{v1} Initial Release. Paper accepted to CVPR 2020, the implementation
will soon be made available.

\section{Qualitative Results}
\subsection*{Pseudo-Labels}
\begin{figure}[H]
  \centering
  \includegraphics[width=\linewidth]{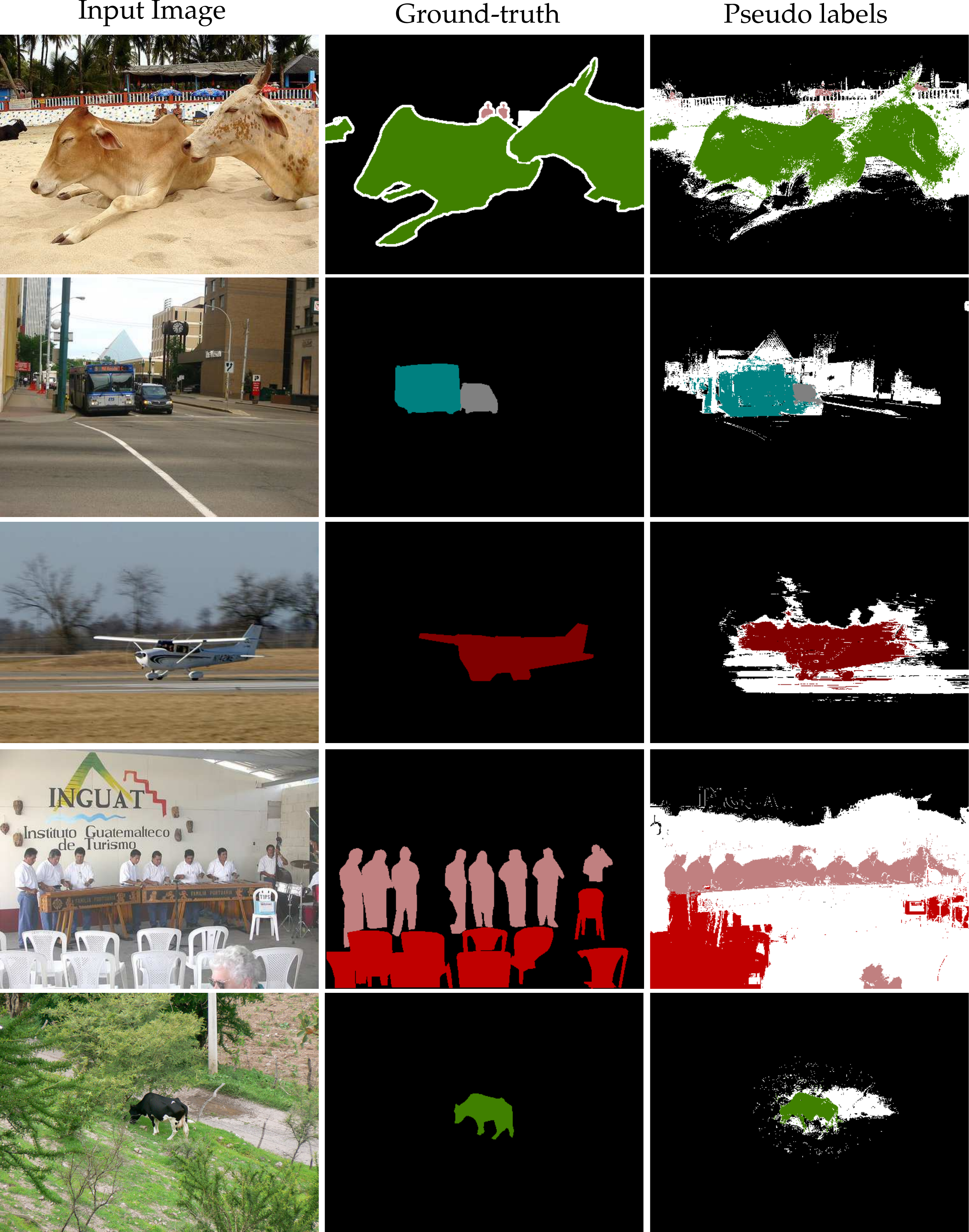}
  \caption{\textbf{The generated pseudo pixel-level labels.}
  Instances of the generated pseudo pixel-level labels
  from PASCAL VOC \textit{train} set.
  The white regions correspond to the ignored pixels.}
  \label{fig:p_labels}
\end{figure}

\cref{fig:p_labels} shows some qualitative results of the generated
pseudo pixel-level labels using the available image-level labels.
We observe that when considering regions with high attention scores (\ie $> 0.3$),
the assigned classes do correspond in most cases to true positives.

\subsection*{Predictions}
\begin{figure*}[ht!]
\centering
  \includegraphics[width=\linewidth]{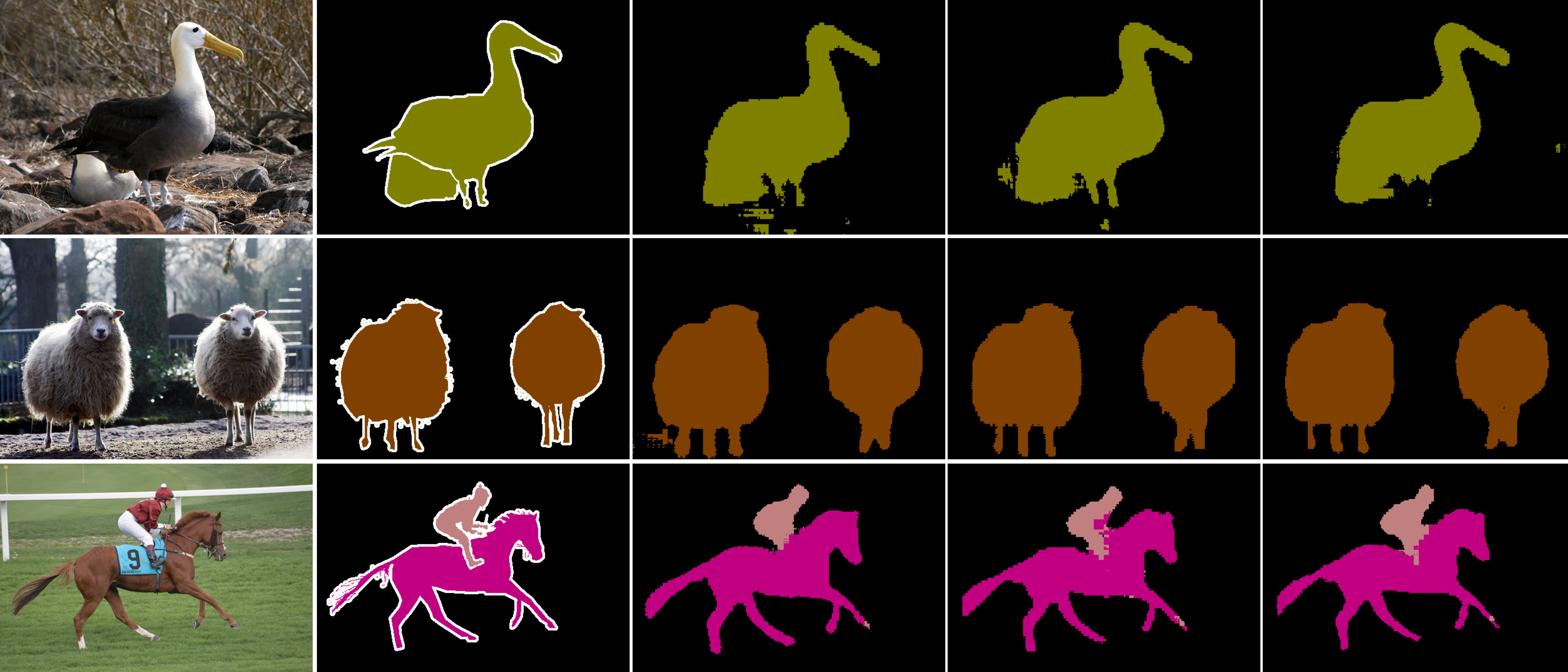}
  \caption{\textbf{CCT results.}
  Semantic Segmentation Results on the PASCAL VOC \textit{val} images.}
  \label{fig:results}
  \vspace{5in}
\end{figure*}

Qualitative results of CCT on PASCAL VOC \textit{val} images
with different values of $n$ are presented in \cref{fig:results}.
\begin{figure*}[ht!]
  \centering
  \includegraphics[width=\linewidth]{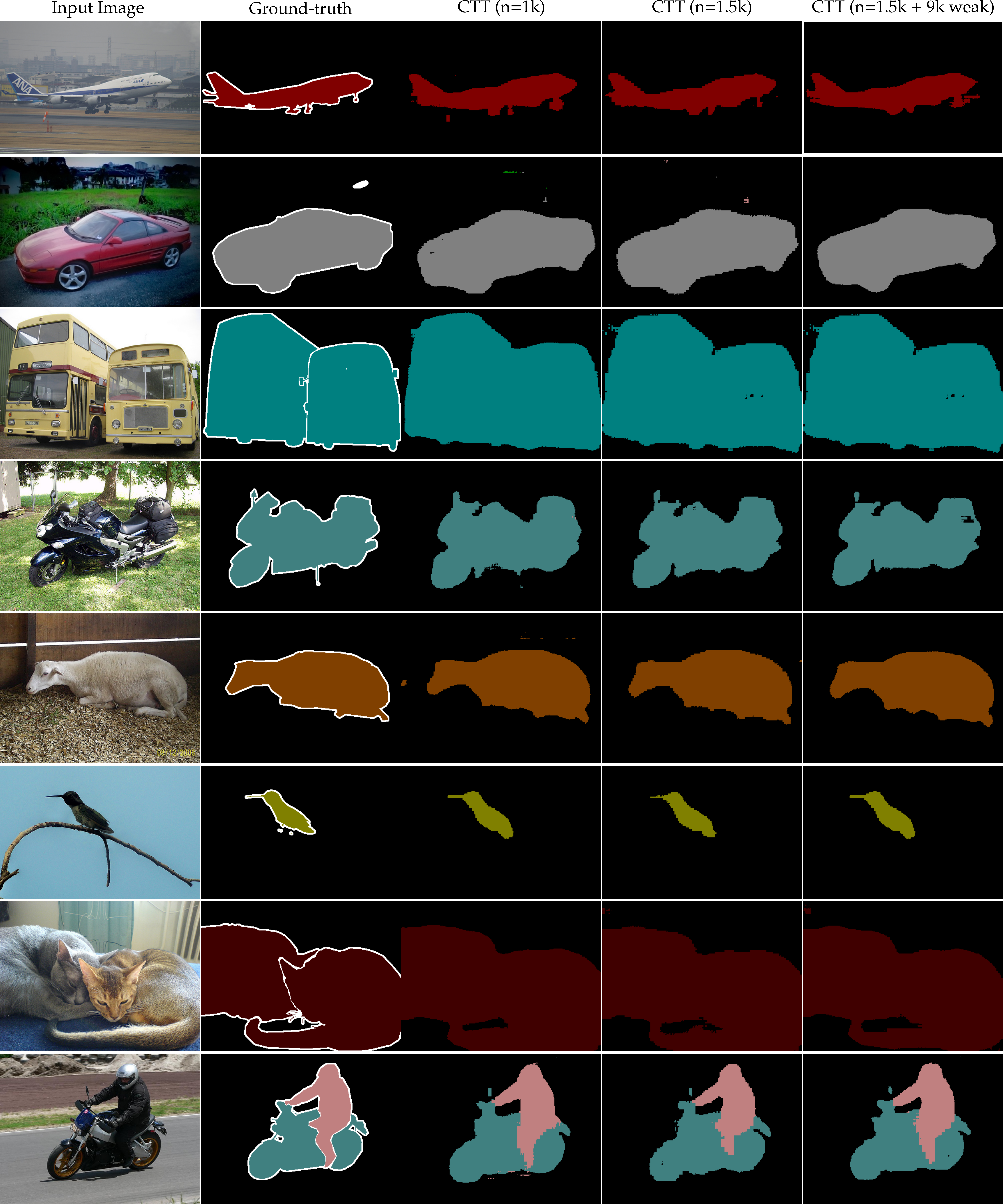}
  \vspace{-0.15in}
\end{figure*}

\end{document}